\newif\ifreview
\reviewfalse 

\newif\ifarxiv
\arxivtrue 

\newif\ifsubmission
\ifreview\submissiontrue\else\submissionfalse\fi

\documentclass[10pt,twocolumn,letterpaper]{article}

\ifreview
  \usepackage[review]{cvpr}
\else
  \ifarxiv
    \usepackage[pagenumbers]{cvpr}
  \else
    \usepackage{cvpr}
  \fi
\fi

\ifsubmission
\PassOptionsToPackage{hypertexnames=false}{hyperref}
\fi
\definecolor{cvprblue}{rgb}{0.21,0.49,0.74}
\usepackage[pagebackref,breaklinks,colorlinks,allcolors=cvprblue]{hyperref}

\usepackage[capitalize]{cleveref}
\usepackage{caption}
\usepackage{subcaption}
\usepackage{graphicx}
\usepackage{hyperref}

\usepackage{lipsum}
\usepackage{wrapfig}
\usepackage{array}
\usepackage{xcolor}
\usepackage{makecell}
\usepackage{enumitem}

\renewcommand{\footnotesize}{\scriptsize}
\usepackage{algorithm}
\usepackage{algpseudocode}
\floatname{algorithm}{Algorithm}
\algrenewcommand\algorithmicrequire{\textbf{Require:}}
\algrenewcommand\algorithmicensure

\usepackage{colortbl}
\usepackage{multirow}
\usepackage{multicol}
\usepackage{listings} 
\usepackage{titletoc} 
\usepackage{fancyvrb}
\usepackage{tcolorbox} 
\usepackage[accsupp]{axessibility}  
\usepackage[subtle]{savetrees} 

\newcommand{\subpara}[1]{%
  \vspace{1mm}%
  \noindent\textbf{#1}%
}

\definecolor{lightgray}{rgb}{0.83, 0.83, 0.83}
\definecolor{Gray}{gray}{0.6}
\definecolor{aliceblue}{rgb}{0.94, 0.97, 1.0}
\definecolor{mistyrose}{rgb}{1.0, 0.89, 0.88}
\definecolor{backcolour}{rgb}{0.95,0.95,0.92}

\crefname{equation}{Eq.}{Eqs.}
\Crefname{equation}{Equation}{Equations}

\crefname{figure}{Fig.}{Figs.}
\Crefname{figure}{Figure}{Figures}

\crefname{table}{Tab.}{Tabs.}
\Crefname{table}{Table}{Tables}

\crefname{section}{Sec.}{Secs.}
\Crefname{section}{Section}{Sections}

\crefname{algorithm}{Alg.}{Algs.}
\Crefname{algorithm}{Algorithm}{Algorithms}

\AtBeginDocument{%
  \setlength{\abovedisplayskip}{3pt}%
  \setlength{\belowdisplayskip}{3pt}%
  \setlength{\abovedisplayshortskip}{2pt}%
  \setlength{\belowdisplayshortskip}{2pt}%
}
\ifsubmission
\usepackage{chapterbib}
\fi

\def\confName{CVPR}
\def\confYear{2027}

\title{Out of Sight, Still in Mind: Token Compression for Omni-LLMs\vspace{-15pt}}

\author{
Suho Yoo$^1$ \quad
Youngjoon Jang$^{1, 2}$ \quad
Hyebin Cho$^1$ \quad
Joon Son Chung$^1$ \\[0.3em]
$^1$KAIST \quad$^2$VGG, University of Oxford
\\ \href{https://github.com/suho-yoo/ReMo}{\textcolor{NavyBlue}{https://github.com/suho-yoo/ReMo}}
}

\makeatletter
\newcommand{\savesuppcitationunit}{\ifsubmission\global\let\supp@citationunit\the@ipfilectr\fi}
\newcommand{\restoresuppcitationunit}{\ifsubmission\global\let\the@ipfilectr\supp@citationunit\fi}
\makeatother
\begin{document}
\ifsubmission
  
\maketitle
\begin{abstract}
The goal of this paper is to reduce the input token cost of Omni-modal large language models (Omni-LLMs) at inference time. Omni-LLMs reason jointly over audio, video and text, but the cost of the three streams is highly unbalanced: visual tokens account for the vast majority of the input, and are highly redundant. In this paper, we propose \textbf{\textit{ReMo}}, a training-free framework that compresses visual tokens by \underline{re}distributing their information across \underline{mo}dalities: a visual token is kept only if its information appears nowhere else. ReMo achieves this in two ways: (i)~it aligns audio and video in a shared space via closed-form calibration, and removes visual tokens already explained by the audio or by other visual tokens; and (ii)~it replaces object-level visual tokens, identified by a lightweight object detector, with compact text proxies, short descriptions of each object and its location, conveying the same content in far fewer tokens. On Qwen2.5-Omni at two model scales, ReMo removes 54\% of the input tokens with comparable accuracy. It performs comparably to the full-token model, reaching 101.2\% and 101.4\% of its average accuracy over five audio-visual benchmarks.
\end{abstract}

\section{Introduction}
\label{sec:intro}

Omni-modal large language models (Omni-LLMs) extend LLMs to perceive audio, vision and text jointly, achieving strong results on real-world audio-visual understanding~\cite{xu2025qwen25omnitechnicalreport, tang2025video, xu2025qwen3, team2026qwen3, hurst2024gpt}. This capability comes at a cost: encoding all three modalities inflates the input sequence to tens of thousands of tokens, most of which come from the visual stream. At this length, the quadratic cost of attention becomes the primary bottleneck for both latency and memory. Reducing the token count without sacrificing accuracy is therefore central to deploying Omni-LLMs at scale.

Token compression tailored to Omni-LLMs has only recently emerged, aiming to reduce the large number of audio or visual input tokens before decoding~\cite{jung2026fastav, park2026omnidrop}. Rather than treating each modality in isolation, these methods exploit cross-modal relationships, such as audio-visual similarity or the relevance of audio and visual tokens to the text query, to identify redundant or less relevant input tokens for pruning. Cross-modal cues guide token pruning by distinguishing complementary information from content already captured by another modality.

Existing approaches broadly fall into two categories depending on whether they require training. 
Training-based methods, such as OmniSIFT~\cite{ding2026omnisift}, EchoingPixels~\cite{gong2025echoingpixels} and ContextGuard~\cite{jung2026keep}, learn a dedicated compression component; whilst effective, they must be retrained for each backbone, limiting their applicability. Training-free methods avoid this cost. DASH~\cite{li2026dash} uses audio boundaries to select informative video frames and adaptively allocate the compression budget. OmniZip~\cite{tao2026omnizip} instead explicitly considers cross-modal redundancy, estimating it from the cosine similarity of raw modality embeddings; however, this assumes the two modalities are comparable in their native spaces, when in fact their embeddings are poorly aligned~\cite{liang2022mind, hong2026textme, lee2026lamb, yoo2026omnikvquant}. OmniScope~\cite{su2026omniscope} and OmniSelect~\cite{yang2026omniselect} address this issue by using external pretrained encoders to estimate multimodal relevance, introducing an additional expensive module outside the Omni-LLM.

\begin{figure}[t]
\centering
\includegraphics[width=\linewidth]{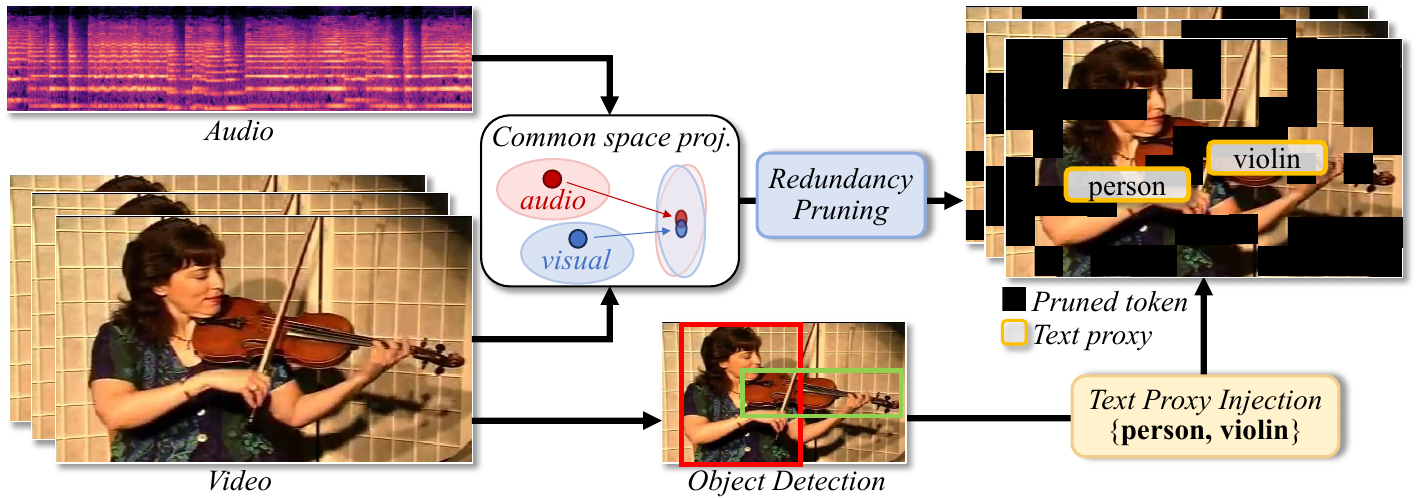}
\vspace{-20pt}
\caption{\textbf{Cross-modal redistribution.} 
A large fraction of visual tokens in an Omni-LLM are redundant. Our method projects audio and video into a common space to drop redundant visual tokens, then replaces the discarded object semantics with compact text proxies, redistributing visual information across the audio and text streams.}
\label{fig:teaser}
\vspace{-15pt}
\end{figure}


In this paper, we revisit how the multimodal capabilities of Omni-LLMs can be used for token compression without fine-tuning. Building on prior use of audio to represent visual semantics, we introduce text as a complementary compression modality, reducing the visual tokens that dominate the input sequence, as illustrated in~\cref{fig:teaser}.
First, ReMo removes only visual tokens whose content is already captured by the accompanying audio, without heavy external modules or gradient updates. Second, pruning visual tokens can discard object-level semantics that are easily expressed in language. We therefore use a lightweight object detector to identify such semantics and insert the detected object names as compact text proxies into the multimodal input. In this way, a few text tokens can preserve information that would otherwise require many visual tokens, while the remaining visual tokens retain information that cannot be supplied by the other modalities.

Each route poses its own challenge. For the first, deciding whether an audio token and a visual token describe the same concept is difficult, because their embeddings live in poorly aligned spaces. We therefore project both modalities into a common space using a closed-form calibration step, where their agreement can be measured directly. For the second, the inserted words must not float free of the scene. We place each description according to the spatial position of the object it replaces, so the text stays anchored to where its content appears. The result, \textbf{\emph{ReMo}}, offloads high-level visual semantics onto the audio and text streams, while reserving visual tokens for complementary visual detail.

Through extensive experiments, we demonstrate that this redistribution works in practice. 
A~controlled audio-to-video retrieval analysis first verifies the projection at the heart of ReMo, improving R@1 from 2.7 to 31.3 over raw embeddings and confirming that audio and video are aligned reliably enough to ground our cross-modal saliency. We then evaluate ReMo on Qwen2.5-Omni~\cite{xu2025qwen25omnitechnicalreport} across four audio-visual understanding benchmarks and one captioning benchmark. At a compression rate comparable to the strongest training-free baseline, ReMo largely preserves accuracy and performs comparably to the full-token model, reaching 101.2\%
and 101.4\% of its average accuracy at the 3B and 7B model sizes. Consistent gains at both sizes, across benchmarks of different natures, and without fine-tuning, suggest that using both audio and text to preserve visual semantics is an effective strategy for compressing Omni-LLM inputs.

\section{Related Work}
\label{sec:related}

\subpara{Omni-modal LLMs.}
Building on multimodal LLMs (MLLMs), a recent line of work extends vision-language models to jointly process text, image, audio, and video within a single LLM, often referred to as Omni-LLMs. Proprietary systems such as GPT-4o~\cite{hurst2024gpt} and Gemini~\cite{team2024gemini, comanici2025gemini} show strong audio-visual reasoning, and a growing body of open models follows~\cite{xu2025qwen25omnitechnicalreport,deshmukh2026nemotron,li2025baichuan, fu2024vita, cheng2024videollama, tang2025video}. These models pair modality-specific encoders with a shared LLM decoder, project each modality into a common embedding space, and interleave the resulting tokens into chunk-structured sequences. A complementary line of work improves these models at inference time through decoding interventions~\cite{jung2026avcd, yoo2026nature, chung2026mad, jung2026probing}. However, such methods change how existing tokens are read while leaving the token budget untouched, so the quadratic cost of attention over thousands of multimodal tokens remains the main bottleneck.


\subpara{Token compression for Omni-LLMs.} Token compression reduces this cost by shrinking the decoder's input sequence. For general multimodal LLMs, methods compress the visual stream by pruning low-attention tokens after an early layer~\cite{chen2024image}, fitting the attention distribution~\cite{ye2025fit}, merging similar tokens~\cite{bolya2022token}, or selecting dominant ones~\cite{visionzip}, without considering audio. Recent Omni-LLM compression methods fall into two groups. Training-free methods include FastAV~\cite{jung2026fastav} and OmniDrop~\cite{park2026omnidrop}, which prune progressively inside the decoder using attention, and OmniZip~\cite{tao2026omnizip}, DASH~\cite{li2026dash}, OmniScope~\cite{su2026omniscope}, and OmniSelect~\cite{yang2026omniselect}, which prune audio-visual tokens at the input level. Training-based methods include OmniSIFT~\cite{ding2026omnisift} and EchoingPixels~\cite{gong2025echoingpixels}, which learn dedicated compression modules, and ContextGuard~\cite{jung2026keep}, which learns a predictor of audio-explainable visual semantics. ReMo performs training-free, input-level compression. While most methods compress both streams, ContextGuard preserves audio and prunes audio-explainable visual tokens. ReMo extends this principle to text, injecting compact proxies to preserve object-level visual semantics while retaining all audio tokens.

\subpara{The modality gap in multimodal models.}
Embeddings from different modalities rarely share a common geometry. Even when encoders are trained to align them, they occupy separate regions of the representation space, a phenomenon characterized as the modality gap~\cite{liang2022mind} and linked to a conical structure in the embedding space. The gap is not specific to one modality pair: it has been reported across image, video, audio, and 3D data~\cite{hong2026textme}, and traced to modality-specific components separable from the shared bimodal signal~\cite{dhimoila2026cross}. A growing line of work seeks to reduce it, either through learned alignment objectives~\cite{lee2026lamb,grassucci2026closing} or by exploiting the geometry of the gap to transfer across modalities without paired supervision~\cite{hong2026textme}. This gap has direct consequences for input-level compression in Omni-LLMs, where audio-video redundancy must be judged before the decoder. OmniZip~\cite{tao2026omnizip} scores this redundancy from raw-embedding cosine similarity, which is directly exposed to the gap. ContextGuard~\cite{jung2026keep} instead trains an audio-to-video prediction module to bridge the two spaces, while OmniScope~\cite{su2026omniscope} and OmniSelect~\cite{yang2026omniselect} rely on pretrained CLIP~\cite{radford2021learning} and AudioCLIP~\cite{guzhov2022audioclip} encoders, respectively. OmniDrop~\cite{park2026omnidrop} sidesteps the problem entirely by pruning in intermediate decoder layers rather than at the input.
In contrast, ReMo uses a simple closed-form calibration to align the two modalities, requiring no gradients or optimizer.


\subpara{Bridging vision and text tokens.}
A complementary line connects visual content with the text token space. ReVisiT~\cite{cho2025revisit} shows that vision tokens intrinsically encode object-level semantics that surface when projected into the text vocabulary, indicating that much of their content is recoverable in language form. ViKey~\cite{lee2026vikey} overlays frame-index prompts on sampled frames and maps query keywords to those indices, using text cues as temporal anchors for sparsely sampled video. In the reverse direction, VIST~\cite{xing2026vision} renders text into images and compresses the rendered features into a few visual tokens, exploiting the spatial redundancy of text. Together these results suggest that object-level semantics transfer between the two token types. In this paper, we exploit this property by replacing redundant visual tokens with compact, spatially grounded text proxies rather than simply discarding them.

\begin{figure*}[t]
    \centering
    \includegraphics[width=\textwidth]{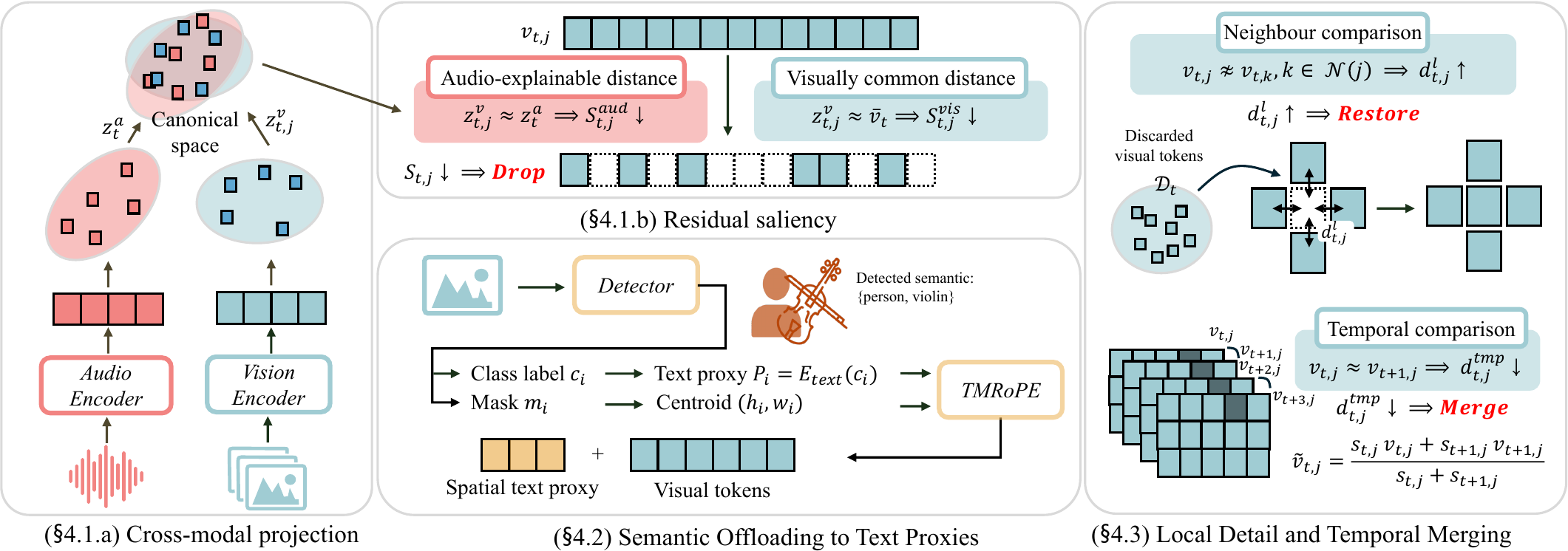}
    \vspace{-20pt}
    \caption{\textbf{Overview of ReMo.}
    ReMo prunes video tokens before the LLM decoder while preserving broad audio--visual context.
    Cross-modal projection (\S4.1) aligns audio tokens $z_t^a$ and video tokens $z_{t,j}^v$ in a canonical space;
    residual saliency drops tokens explained by audio ($S^{aud}$) or common visual context ($S^{vis}$), keeping audio- and video-unexplained evidence such as rare visual details, while semantic offloading (\S4.2) routes detected objects to text proxies $P_i$ whose positions are preserved via TMRoPE;
    local detail restoration (\S4.3) recovers high-neighbour-distance tokens $d^l_{t,j}$ from the discarded set $\mathcal{D}_t$, and temporal merging fuses redundant co-located tokens.}
    \label{fig:main}
\vspace{-5pt}
\end{figure*}

\section{Preliminaries}
\label{sec:prelim}

\subpara{Background on Omni-LLMs.}
We build on Qwen2.5-Omni~\cite{xu2025qwen25omnitechnicalreport}, which reads
text, audio, and video within a single sequence.
Text is tokenized with a byte-level BPE tokenizer~\cite{sennrich2016neural, radford2019language}, audio is resampled and turned
into mel-spectrogram features whose encoder emits one token per short fixed
audio window, and video frames are encoded by a vision transformer into patch
tokens.
For a video with its audio track, the two streams are interleaved at the chunk
level rather than token by token.
The timeline is divided into $T$ successive temporal chunks of a fixed duration,
and each chunk contributes its visual tokens first and its audio tokens after
them, so the full input is a concatenation of per-chunk blocks,
\begin{equation}
X=\big[\,V_1, A_1,\; V_2, A_2,\; \dots,\; V_T, A_T\,\big],
\label{eq:interleave}
\end{equation}
where $V_t=\{v_{t,j}\}_{j=1}^{N}$ and $A_t=\{a_{t,i}\}_{i=1}^{M}$ are the sets of $N$ visual tokens and $M$ audio tokens in chunk $t$, respectively. This layout orders the sequence by time while keeping the visual and audio tokens of each interval adjacent.
Because each frame yields a full patch grid, the visual blocks $V_t$ dominate the sequence and drive the quadratic attention cost our method targets.

\subpara{Temporal-spatial positional encoding.}
Token positions follow TMRoPE~\cite{xu2025qwen25omnitechnicalreport}, which
extends M-RoPE~\cite{wang2024qwen2} by adding absolute time to rotary
embeddings~\cite{su2024roformer}. For a one-dimensional position $p$, RoPE
rotates each feature pair according to
\begin{equation}
R_\Theta(p)=\bigoplus_{j=1}^{d/2}
\begin{pmatrix}
\cos p\theta_j & -\sin p\theta_j\\[2pt]
\sin p\theta_j & \cos p\theta_j
\end{pmatrix},
\quad
\theta_j=b^{-2j/d}.
\label{eq:rope}
\end{equation}
TMRoPE extends this to three axes by assigning each token a three-part position
$(t,h,w)$ over the temporal, height, and width,
\begin{equation}
R_{(t,h,w)}
=
\mathrm{diag}\big(R_t(t),R_h(h),R_w(w)\big),
\label{eq:tmrope}
\end{equation}
where $R_t$, $R_h$, and $R_w$ are instances of the rotary embedding in
Eq.~\eqref{eq:rope} applied to the temporal, height, and width axes,
respectively. During attention, these rotations are applied to the query and
key representations based on the position of each token, preserving relative
relationships across the three axes.
Text tokens share one index across all axes, audio tokens advance only on the
temporal axis, and video tokens take spatial indices from the patch grid while
their temporal index tracks real time, so the visual and audio tokens of a
chunk land at matching times.
This later allows us to assign text proxies to the temporal and spatial
positions of the visual objects they replace.
\section{Method}


\subpara{Overview.}
As illustrated in~\cref{fig:main}, ReMo is a gradient-free method that treats compression as redistribution rather than removal: a visual token is kept only when its information is irrecoverable elsewhere.
Operating on the input token sequence of the LLM,
it first projects audio and visual tokens into a common embedding space to remove visual tokens already explained by audio or other visual tokens (Sec.~\ref{sec:proj}),
injects text proxies for object-level meaning otherwise lost while preserving their temporal and spatial positions (Sec.~\ref{sec:text}),
and restores local visual details and merges temporally redundant tokens (Sec.~\ref{sec:detail}).

\subsection{Scoring Visual Redundancy}
\label{sec:proj}
We first remove visual tokens whose content is already carried by the audio.
The natural way to measure this overlap is cosine similarity between audio and
visual embeddings, but the modality gap places the two in separate regions of
the embedding space, making raw similarity unreliable.
Instead of training an alignment module, we design a simple yet effective
closed-form map that aligns the two modalities in a common space.

\subpara{Cross-modal projection.}
Given paired audio--visual embeddings $(\mathbf{A},\mathbf{V})$ from
VGGSound~\cite{chen2020vggsound}, we centre them
($\Delta\mathbf{A}=\mathbf{A}-\mu_a$, $\Delta\mathbf{V}=\mathbf{V}-\mu_v$) and
whiten each by the regularized inverse square root of its covariance,
\begin{equation}
\begin{aligned}
W_a &= \Big(\tfrac1n\Delta\mathbf{A}^\top\Delta\mathbf{A}+\epsilon I\Big)^{-1/2},\\
W_v &= \Big(\tfrac1n\Delta\mathbf{V}^\top\Delta\mathbf{V}+\epsilon I\Big)^{-1/2}.
\end{aligned}
\end{equation}
The SVD of the whitened cross-covariance in regularized CCA,
\begin{equation}
C=\tfrac1n\,(\Delta\mathbf{A}\,W_a)^\top(\Delta\mathbf{V}\,W_v)=U\Sigma Q^\top,
\end{equation}
yields singular values $\Sigma=\mathrm{diag}(\sigma_1,\dots,\sigma_d)$, the
canonical correlations between the modalities. Keeping the top-$K$ directions
scaled by $\sigma_k^{1/2}$ gives
\begin{equation}
P_a=W_a\,U_K\,\Sigma_K^{1/2},
\qquad
P_v=W_v\,Q_K\,\Sigma_K^{1/2}.
\label{eq:proj}
\end{equation}
At inference, we map the mean audio representation $\bar a_t$ and visual tokens into the common space as $z^a_t=(\bar a_t-\mu_a)P_a$ and
$z^v_{t,j}=(v_{t,j}-\mu_v)P_v$, with $z^a_t$ serving as an audio anchor.

\subpara{Residual saliency.}
A token is worth keeping only if informative in both senses, neither explained by its chunk's pattern nor recoverable from audio,
\begin{equation}
\begin{aligned}
s^{\mathrm{vis}}_{t,j} &= 1-\cos\big(v_{t,j},\bar v_t\big),\\
s^{\mathrm{aud}}_{t,j} &= 1-\cos\big(z^v_{t,j},z^a_t\big),
\end{aligned}
\label{eq:saliency_terms}
\end{equation}
with $\bar v_t$ the chunk mean. Since the two differ in scale, we fuse them with a per-chunk weight,
\begin{equation}
s_{t,j}=(1-\beta_t)\,s^{\mathrm{vis}}_{t,j}+\beta_t\,s^{\mathrm{aud}}_{t,j},
\qquad
\beta_t=\frac{\bar s^{\mathrm{vis}}_t}{\bar s^{\mathrm{vis}}_t+\bar s^{\mathrm{aud}}_t},
\label{eq:saliency}
\end{equation}
where $\bar s^{(\cdot)}_t$ is the chunk-average magnitude. Given a budget
$\rho_{av}$, we keep the $\lceil\rho_{av} N\rceil$ highest-scoring tokens as
$\mathcal{S}_t$, dropping first those both visually common and predictable from
audio.

\subsection{Semantic Offloading to Text Proxies}
\label{sec:text}
Saliency pruning removes redundant tokens, but the object-level semantics they
carried would vanish with them, though such semantics cost far fewer words than
pixels. Rather than spend visual tokens on them, we name the objects explicitly.
A lightweight detector~\cite{yaseen2024yolov8indepthexplorationinternal} runs on
the frames $I_t$ of chunk $t$ and returns $M_t$ detections,
\begin{equation}
\{(c_i,m_i)\}_{i=1}^{M_t}=\mathrm{Det}(I_t),
\label{eq:detect}
\end{equation}
each pairing an object class $c_i$ with its region mask $m_i$. We turn every
class into a text proxy through the model's text embedding,
$p_i=E_{\mathrm{text}}(c_i)$.

To spatially ground each proxy, we assign it a TMRoPE position. Unlike text
tokens, which normally share one index across all three axes, we keep the
temporal index at chunk $t$ and obtain the spatial indices from the region
centroid, $(h_i,w_i)=\mathrm{centroid}(m_i)$, assigning
\begin{equation}
\operatorname{pos}(p_i)=(t,h_i,w_i).
\label{eq:place}
\end{equation}
During attention, TMRoPE applies this position to the query and key representations of $p_i$, allowing the retained visual tokens to attend to the proxy as if it were placed at the object's temporal and spatial location, encoding both \emph{what} the object is and \emph{where} it lies. We prepend these proxies to the retained visual tokens of their chunk, so that each chunk reads as a compact summary followed by the residual visual detail it could not express.

\subsection{Local Detail and Temporal Merging}
\label{sec:detail}
Two effects remain. Cosine saliency favours global similarity and can suppress
local visual structure, while adjacent chunks often repeat surviving content.
We address both after selection.

\subpara{Local detail.}
For each token we measure its deviation from its four spatial neighbours $\mathcal{N}(j)$,
\begin{equation}
d^l_{t,j}=\tfrac{1}{|\mathcal{N}(j)|}\sum_{k\in\mathcal{N}(j)}\lVert v_{t,j}-v_{t,k}\rVert_2 ,
\label{eq:local}
\end{equation}
which favours local structure rather than semantic similarity and thus
complements saliency. Among the discarded tokens $\mathcal{D}_t$, we restore the
$\lceil\rho_l N\rceil$ with the highest $d^l_{t,j}$.

\subpara{Temporal merging.}
We slide a two-chunk window across the sequence. For co-located tokens in chunks
$t$ and $t{+}1$, we merge a pair when they are similar
($d^{tmp}_{t,j}=1-\cos(v_{t,j},v_{t+1,j})<\tau$) and the pair's minimum saliency
$m_j=\min(s_{t,j},s_{t+1,j})$ falls below median of $\{m_j\}$ over the
window, by saliency-weighted fusion
\begin{equation}
\tilde v_{t,j}=\frac{s_{t,j}\,v_{t,j}+s_{t+1,j}\,v_{t+1,j}}{s_{t,j}+s_{t+1,j}},
\label{eq:temporal}
\end{equation}
biasing the fused token toward the one with more residual information and
inheriting its position.
\section{Experiments}
\begin{table*}[t]
\centering
\captionsetup{width=1\textwidth}
\caption{\textbf{Main results on Qwen2.5-Omni at the 3B and 7B scales.}
Compared with the strongest training-free baseline, ReMo largely preserves accuracy and performs comparably to the full-token model. \emph{Comp.} is the average token
compression rate measured over the full input sequence,
and \emph{Avg.} normalizes each benchmark to the full-token score before averaging.
Full-token rows are shown in \textcolor{gray}{gray} as reference.
Among compression methods, best and second-best benchmark scores
and normalized averages are shown in \textbf{bold} and
\underline{underlined}, respectively.}
\vspace{-10pt}
\label{tab:main}
\setlength{\tabcolsep}{8pt}
\scriptsize
\resizebox{\linewidth}{!}{%
\begin{tabular}{lccccccc}
\toprule
Method & Comp.$\uparrow$(\%) & World.$\uparrow$ & Daily.$\uparrow$ &
MME$\uparrow$ & OmniVid.$\uparrow$ &
video-SAL2.$\downarrow$ & Avg.$\uparrow$(\%) \\
\midrule
\multicolumn{8}{c}{\textit{Qwen2.5-Omni$_{3\mathrm{B}}$}} \\
\midrule
\rowcolor{gray!8}
Full Token & 0 & 47.7 & 57.7 & 75.8 & 44.0 & 53.5 & 100.0 \\
Random & 50 & 44.1 & 53.1 & 74.0 & 42.8 & 56.1 & 95.0 \\
FastV (ECCV'24) & 50 & 46.7 & 55.6 & 74.0 & \underline{44.0} & 54.6 & 98.0 \\
OmniZip (CVPR'26) & 54 & 47.1 & 55.8 & 74.9 & 42.8 & \textbf{52.4} & 98.7 \\
DASH (ECCV'26) & 55 & \underline{47.3} & 55.7 & \underline{76.2} & \textbf{44.6} & 53.9 & \underline{99.4} \\
OmniScope (ACM MM'26) & 55 & 46.9 & \underline{57.5} & 72.7 & 42.2 & 53.9 & 97.8 \\
OmniSelect (NeurIPS'26) & 54 & 47.1 & 55.7 & \underline{76.2} & 42.2 & 56.2 & 97.4 \\
\rowcolor{aliceblue}
\textbf{Ours} & 54 & \textbf{47.8} & \textbf{58.3} & \textbf{77.5} & \textbf{44.6} & \underline{52.9} & \textbf{101.2} \\
\midrule
\multicolumn{8}{c}{\textit{Qwen2.5-Omni$_{7\mathrm{B}}$}} \\
\midrule
\rowcolor{gray!8}
Full Token & 0 & 47.4 & 57.0 & 78.8 & 47.6 & 49.3 & 100.0 \\
Random & 50 & 45.7 & 52.4 & \underline{77.5} & 43.4 & 50.4 & 95.1 \\
FastV (ECCV'24) & 50 & 45.6 & \underline{56.6} & \textbf{77.9} & \underline{47.6} & 49.8 & 98.7 \\
OmniZip (CVPR'26) & 54 & 46.8 & \underline{56.6} & 77.1 & 47.0 & 52.8 & 97.6 \\
DASH (ECCV'26) & 55 & 46.6 & 51.6 & 75.8 & \underline{47.6} & \underline{48.8} & 97.2 \\
OmniScope (ACM MM'26) & 55 & \underline{47.3} & 56.4 & 75.8 & \textbf{49.4} & 49.4 & \underline{99.7} \\
OmniSelect (NeurIPS'26) & 54 & 46.4 & 54.8 & 76.6 & 44.0 & 51.6 & 95.8 \\
\rowcolor{aliceblue}
\textbf{Ours} & 54 & \textbf{48.2} & \textbf{57.4} & \textbf{77.9} & \textbf{49.4} & \textbf{48.4} & \textbf{101.4} \\
\bottomrule
\end{tabular}%
}
\vspace{-10pt}
\end{table*}


\subsection{Experimental Setup}
\subpara{Datasets and metrics.}
We evaluate on four audio-visual understanding benchmarks, WorldSense~\cite{hong2025worldsense} (World.), DailyOmni~\cite{zhou2025daily} (Daily.), Video-MME~\cite{fu2025video} (MME), and OmniVideoBench~\cite{li2025omnivideobench} (OmniVid.), and one captioning benchmark, video-SALMONN2 test set~\cite{tang2025video} (video-SAL2.). For long-video benchmarks we evaluate only QA pairs whose videos are under one minute, as full-token inference on the full sets exceeds our memory. We report per-benchmark accuracy and an aggregate \emph{Avg.} that normalizes each benchmark so the full-token model scores $100$ and averages across benchmarks. For video-SAL2., where lower is better, we use the inverse ratio. We also report \emph{Comp.}, the average compression rate relative to the full input sequence.

\subpara{Implementation details.}
We apply ReMo to Qwen2.5-Omni (3B/7B), and all experiments run on a single NVIDIA RTX A6000 (48GB) GPU. The cross-modal projection is built once and offline from $5{,}000$ paired
audio-visual embeddings on VGGSound train set, keeping the top $K{=}512$ canonical
directions. We set the selection budget $\rho_{av}{=}0.4$, the local recovery
budget $\rho_l{=}0.05$, and the temporal-merging threshold $\tau{=}0.05$, fixed
across all benchmarks. Object classes for text proxies are detected with a
lightweight YOLO~\cite{yaseen2024yolov8indepthexplorationinternal} detector.

\subpara{Baselines.}
We compare against training-free token compression methods. Random drops visual and audio tokens uniformly at random. FastV~\cite{chen2024image} prunes by attention, which we extend to all audio-visual tokens. OmniZip~\cite{tao2026omnizip} prunes at the input level by selecting salient audio tokens and removing video tokens by their raw-embedding cosine similarity to the retained audio.
DASH~\cite{li2026dash} uses changes in the audio stream to guide which parts of the video receive more or fewer tokens.
OmniScope~\cite{su2026omniscope} uses the text query to score audio and video separately and allocates modality-specific token budgets.
OmniSelect~\cite{yang2026omniselect} uses AudioCLIP~\cite{guzhov2022audioclip} to determine whether each input is audio- or video-centric and adjusts the pruning ratio accordingly. Full Token denotes the uncompressed model.
Hyperparameter settings for all baselines are provided in~\cref{app:baseline_settings} of supp.mat.

\subsection{Experimental Results}

\cref{tab:main} reports results across the four audio-visual benchmarks and captioning.
At compression rates comparable to the training-free baselines, ReMo performs comparably to the full-token model, reaching $101.2\%$ and $101.4\%$ of its average accuracy on the 3B and 7B backbones.
In contrast, all competing methods remain below the full-token model, with the strongest baselines reaching $99.4\%$ on 3B (DASH) and $99.7\%$ on 7B (OmniScope).
ReMo achieves the best score among compression methods on all four audio-visual benchmarks at both scales, with lower captioning error than Full Token on video-SAL2.
The largest gains appear on MME for 3B ($75.8 \rightarrow 77.5$) and OmniVid.\ for 7B ($47.6 \rightarrow 49.4$).
This shows that ReMo recovers what pruning would otherwise discard.

\subsection{Analysis}

\begin{table}[t]
\centering
\caption{\textbf{Component ablation.} Each row removes a single component (text proxies, local detail, text position, projection, or temporal merging) on the 3B model.
The two recovery mechanisms matter most. Removing them raises
compression but loses the most accuracy, while the remaining components only refine the result at little cost.}
\vspace{-0.48em}
\label{tab:ablation}
\setlength{\tabcolsep}{7pt}
\scriptsize
\resizebox{\linewidth}{!}{%
\begin{tabular}{lccc}
\toprule
Method & World.$\uparrow$ & Daily.$\uparrow$ & Comp.$\uparrow$ \\
\midrule
\rowcolor{aliceblue}
\textbf{Ours} & \textbf{47.8} & \textbf{58.3} & 54 \\
\midrule
\quad w/o local    & 46.4\,\textcolor{red}{\scriptsize$(-1.4)$} & 57.5\,\textcolor{red}{\scriptsize$(-0.8)$} & 56 \\
\quad w/o text     & 46.7\,\textcolor{red}{\scriptsize$(-1.1)$} & 57.1\,\textcolor{red}{\scriptsize$(-1.2)$} & 55 \\
\quad w/o rope     & 47.2\,\textcolor{red}{\scriptsize$(-0.6)$} & 58.0\,\textcolor{red}{\scriptsize$(-0.3)$} & 54 \\
\quad w/o proj.    & 47.5\,\textcolor{red}{\scriptsize$(-0.3)$} & 58.1\,\textcolor{red}{\scriptsize$(-0.2)$} & 54 \\
\quad w/o temporal & 47.5\,\textcolor{red}{\scriptsize$(-0.3)$} & 58.1\,\textcolor{red}{\scriptsize$(-0.2)$} & 52 \\
\bottomrule
\end{tabular}%
}
\vspace{-15pt}
\end{table}
\subpara{Component ablation.}
\cref{tab:ablation} removes one component at a time. Every component helps, but
the two recovery mechanisms matter most. Dropping local-detail recovery (\emph{w/o
local}, \cref{eq:local}) is the single largest hit on World. ($-1.4$), where
fine spatial cues such as edges and texture matter. Removing the text proxies (\emph{w/o
text}, \cref{sec:text}) costs $1.1$ and $1.2$ points on World. and Daily. while
pushing compression up to $55\%$, confirming that the proxies carry real
semantic content rather than padding. Removing the proxy position
(\emph{w/o rope}, using $(t,t,t)$ instead of $(t,h,w)$, \cref{eq:place}) or the projection to
common space (\emph{w/o proj.}, \cref{eq:proj}) yields negligible cost but still affects
accuracy. Temporal merging (\emph{w/o temporal}, \cref{eq:temporal}) trades the
other way, adding $2\%$ compression at less accuracy cost.

\begin{table*}[t]
\centering
\captionsetup{width=1\textwidth}
\caption{\textbf{Results on Qwen3-Omni-30B (Instruct and Thinking).} ReMo
transfers to a much larger backbone without any change to the method, maintaining comparable average performance on both variants under a matched compression rate.}
\vspace{-10pt}
\label{tab:main_q3}
\setlength{\tabcolsep}{14pt}
\scriptsize
\resizebox{\linewidth}{!}{%
\begin{tabular}{lcccccc}
\toprule
Method & Comp.$\uparrow$(\%) & World.$\uparrow$ & Daily.$\uparrow$ &
MME$\uparrow$ & OmniVid.$\uparrow$ & Avg.$\uparrow$(\%) \\
\midrule
\multicolumn{7}{c}{\textit{Qwen3-Omni-30B$_{\mathrm{Instruct}}$}}\\
\midrule
\rowcolor{gray!8} Full Token & 0 & 53.6 & 69.3 & 84.8 & 48.2 & 100.0 \\
Random   & 50 & 52.2 & 66.6 & 84.0 & 47.0 & 97.5 \\
OmniZip (CVPR’26)  & 58 & 53.2 & 67.6 & 83.5 & 48.8 & 99.1 \\
\rowcolor{aliceblue}
\textbf{Ours} & 57 & \textbf{53.3} & \textbf{67.7} & \textbf{84.4} & \textbf{51.2} & \textbf{100.7} \\
\midrule
\multicolumn{7}{c}{\textit{Qwen3-Omni-30B$_{\mathrm{Thinking}}$}} \\
\midrule
\rowcolor{gray!8} Full Token & 0 & 53.9 & 63.5 & 84.4 & 48.2 & 100.0 \\
Random   & 50 & 49.6 & 62.2 & 81.0 & 44.0 & 94.3 \\
OmniZip (CVPR’26)  & 58 & 54.4 & 63.1 & 82.7 & 47.6 & 99.3 \\
\rowcolor{aliceblue}
\textbf{Ours} & 57 & \textbf{54.6} & \textbf{63.8} & \textbf{83.5} & \textbf{48.2} & \textbf{100.2} \\
\bottomrule
\end{tabular}%
}
\vspace{-10pt}
\end{table*}
\subpara{Scaling to a larger backbone.}
\cref{tab:main_q3} applies ReMo to Qwen3-Omni-30B-A3B~\cite{xu2025qwen3}
in both its Instruct and Thinking variants. Removing $57\%$ of the input
tokens, ReMo maintains average accuracy comparable to the full-token model,
reaching $100.7\%$ and $100.2\%$ on Instruct and Thinking, respectively.
It also outperforms OmniZip on both variants at a comparable compression
rate.
This tenfold increase in backbone size, from 3B to 30B, supports the effectiveness of our method at larger model scales.
Experimental details are provided in~\cref{app:q3} of supp.mat.

\begin{table}[t]
\centering
\caption{\textbf{Efficiency comparison.} ReMo matches the fastest baseline in
latency and memory while far exceeding its accuracy.}
\vspace{-10pt}
\label{tab:efficiency}
\setlength{\tabcolsep}{7pt}
\scriptsize
\resizebox{\linewidth}{!}{%
\begin{tabular}{lccc}
\toprule
\makecell[l]{Method} & \makecell{GPU Mem.$\downarrow$ (GiB)} & \makecell{Latency$\downarrow$ (sec)} & \makecell{Rel.$\downarrow$} \\
\midrule
\multicolumn{4}{c}{\textit{Qwen2.5-Omni$_{3\mathrm{B}}$}} \\
\midrule
\rowcolor{gray!8}
Full Token & 19.0 & 6.9 & 1.00 \\
FastV     & 14.2 & 5.0 & 0.72 \\
OmniZip   & \textbf{13.5} & \textbf{4.6} & \textbf{0.67} \\
DASH      & \underline{13.6} & \underline{4.7} & \underline{0.68} \\
OmniScope & 14.3 & 5.9 & 0.86 \\
OmniSelect& 14.4 & 5.0 & 0.72 \\
\rowcolor{aliceblue}
\textbf{Ours} & \underline{13.6} & \underline{4.7} & \underline{0.68} \\
\midrule
\multicolumn{4}{c}{\textit{Qwen2.5-Omni$_{7\mathrm{B}}$}} \\
\midrule
\rowcolor{gray!8}
Full Token & 27.7 & 9.8 & 1.00 \\
FastV     & 22.6 & 6.7 & 0.68 \\
OmniZip   & \textbf{21.8} & \textbf{5.9} & \textbf{0.60} \\
DASH      & \underline{21.9} & \textbf{5.9} & \textbf{0.60} \\
OmniScope & 22.7 & 7.2 & 0.73 \\
OmniSelect& 22.7 & 6.1 & 0.62 \\
\rowcolor{aliceblue}
\textbf{Ours} & \underline{21.9} & \underline{6.0} & \underline{0.61} \\
\bottomrule
\end{tabular}%
}
\vspace{-15pt}
\end{table}
\subpara{Efficiency.}
\cref{tab:efficiency} compares memory and latency on MME. ReMo remains close to the most efficient baselines, requiring only $13.6/21.9$ GiB of memory and $4.7/6.0$ seconds on the 3B/7B backbones, within $0.1$ GiB and $0.1$ seconds of OmniZip. In contrast, FastV retains a larger memory footprint because pruning occurs only after the decoder has processed the full input at early layers. OmniScope and OmniSelect additionally rely on external CLIP~\cite{radford2021learning} and AudioCLIP~\cite{guzhov2022audioclip} encoders, respectively, and incur higher memory and latency. 
ReMo also uses a lightweight detector for text proxies at inference time, with its small overhead included in the reported memory and latency.
We provide a detailed comparison of the external modules in~\cref{app:external_module} of supp.mat.


\begin{table}[t]
\centering
\caption{\textbf{Audio-to-video retrieval on the VGGSound test set}. We retrieve
the matching video for each audio query, comparing raw embeddings (\emph{Raw}) against our projected
embeddings (\emph{Ours}). The projection improves alignment, confirming that redundancy
is reliably measurable in the common space.}
\vspace{-0.9em}
\label{tab:retrieval}
\setlength{\tabcolsep}{11pt}
\scriptsize
\resizebox{\linewidth}{!}{%
\begin{tabular}{lcccc}
\toprule
Method & R@1$\uparrow$ & R@5$\uparrow$ & R@10$\uparrow$ & MedR$\downarrow$ \\
\midrule
\multicolumn{5}{c}{\textit{Qwen2.5-Omni$_{3\mathrm{B}}$}} \\
\midrule
Raw          &  2.7 & 10.0 & 16.5 & 62 \\
\rowcolor{aliceblue}
\textbf{Ours} & \textbf{31.3} & \textbf{58.0} & \textbf{67.3} & \textbf{4} \\
\midrule
\multicolumn{5}{c}{\textit{Qwen2.5-Omni$_{7\mathrm{B}}$}} \\
\midrule
Raw          &  5.0 & 14.3 & 21.0 & 52 \\
\rowcolor{aliceblue}
\textbf{Ours} & \textbf{35.6} & \textbf{64.0} & \textbf{73.4} & \textbf{3} \\
\bottomrule
\end{tabular}%
}
\vspace{-5pt}
\end{table}

\begin{figure}[t]
\centering
\includegraphics[width=\linewidth]{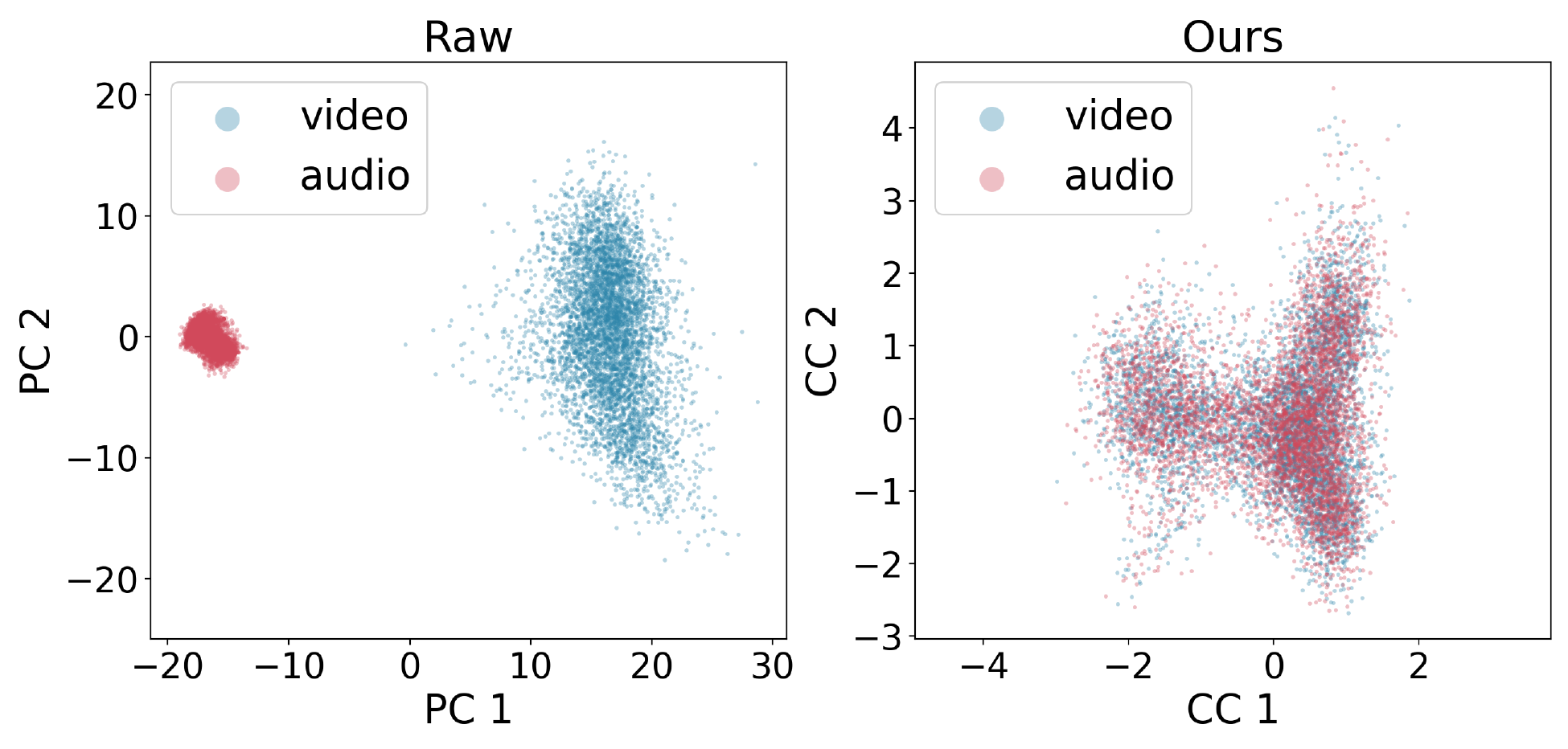}
\vspace{-2em}
\caption{\textbf{Audio and visual embeddings on VGGSound test set.}
\textbf{Left:} raw space by PCA, where the two modalities separate and differ
in spread.
\textbf{Right:} our projection along the top two canonical components. The
paired embeddings now overlap in a common space, making audio-visual similarity
directly measurable.}
\label{fig:pca}
\vspace{-15pt}
\end{figure}

\begin{table*}[t]
\centering
\caption{\textbf{Long-video and streaming evaluation.}
ReMo retains near-full-token average performance at 3B and performs similarly in both settings at 7B
\emph{Comp.} reports the input token compression rate.}
\vspace{-10pt}
\label{tab:long_streaming}
\setlength{\tabcolsep}{8pt}
\small
\resizebox{\linewidth}{!}{%
\begin{tabular}{lcc cc|cccc}
\toprule
& & \multicolumn{3}{c}{\textit{Long-video}}
& \multicolumn{4}{c}{\textit{Streaming}} \\
\cmidrule(lr){3-5}\cmidrule(lr){6-9}
Method & Comp.$\uparrow$(\%)
& \makecell{MME Long$\uparrow$}
& LVOmniBench$\uparrow$ & Avg.$\uparrow$(\%)
& StreamingBench$\uparrow$ & FutureOmni$\uparrow$
& OmniMMI$\uparrow$ & Avg.$\uparrow$(\%) \\
\midrule
\multicolumn{9}{c}{\textit{Qwen2.5-Omni$_{3\mathrm{B}}$}} \\
\midrule
\rowcolor{gray!8}
Full Token & 0
& 48.4 & 32.1 & 100.0
& 38.9 & 45.8 & 33.2 & 100.0 \\
Random & 50
& 47.6 & 31.1 & 97.6
& 37.3 & 46.3 & 31.3 & 97.1 \\
\rowcolor{aliceblue}
\textbf{Ours} & 54
& \textbf{48.9} & \textbf{32.1} & \textbf{100.5}
& \textbf{38.4} & \textbf{47.4} & \textbf{31.8} & \textbf{99.3} \\
\midrule
\multicolumn{9}{c}{\textit{Qwen2.5-Omni$_{7\mathrm{B}}$}} \\
\midrule
\rowcolor{gray!8}
Full Token & 0
& 50.8 & 34.0 & 100.0
& 35.8 & 48.4 & 39.0 & 100.0 \\
Random & 50
& 50.8 & 33.1 & 98.7
& 36.8 & 47.3 & 36.3 & 97.9 \\
\rowcolor{aliceblue}
\textbf{Ours} & 54
& \textbf{51.7} & \textbf{34.3} & \textbf{101.3}
& \textbf{38.1} & \textbf{48.7} & \textbf{39.1} & \textbf{102.4} \\
\bottomrule
\end{tabular}%
}
\vspace{-15pt}
\end{table*}

\subpara{Cross-modal alignment.}
Our saliency (\cref{sec:proj}) assumes that audio and video become comparable
once projected. \cref{fig:pca} shows this qualitatively, in the raw space the
modalities form disjoint clusters, while after projection the paired
embeddings overlap in a common space. We confirm this quantitatively with
audio-to-video retrieval on the VGGSound test set, sampling one audio-video
pair per class and averaging over $50$ sampling seeds (\cref{tab:retrieval}).
In the raw space, retrieval is poor, with R@1 of $2.7$ and $5.0$
and median rank above $50$. After projection, R@1 rises to $31.3$ and
$35.6$ and median rank falls to $4$ and $3$. The projection thus aligns the
modalities well enough for the cross-modal term in our saliency to reflect
genuine semantic overlap, the premise the redundancy scoring rests on.

\subpara{Long-video and streaming evaluation.}
We further evaluate ReMo on Video-MME Long~\cite{fu2025video}
and LVOmniBench~\cite{tao2026lvomnibench} for long-video understanding,
and StreamingBench~\cite{lin2026streamingbench},
FutureOmni~\cite{chen2026futureomni}, and OmniMMI~\cite{wang2025omnimmi}
for streaming evaluation, with results reported in~\cref{tab:long_streaming}.
On the 3B backbone, ReMo retains near full-token performance despite
removing $54\%$ of the input, reaching $100.5\%$ and $99.3\%$ average accuracy in the long-video and streaming settings,
respectively. On 7B, it performs comparably to the full-token model in both settings,
reaching $101.3\%$ and $102.4\%$. These results show that ReMo remains
effective beyond the short-video setting used in~\cref{tab:main}.
Due to GPU memory constraints, experiments are run on four NVIDIA RTX A6000 GPUs; benchmark and
implementation details are provided in~\cref{app:long_streaming}
of the supp. mat.


\begin{table}[t]
\centering
\caption{\textbf{Text proxies help only after pruning}. On the full
video set the text proxies ($N_{add}$ tokens averaged across
World. and Daily.) leave accuracy unchanged since the
information is already present, but on a random 50\% prune they recover much of
the lost accuracy. The proxy helps only once tokens are pruned, not as a generic
add-on.}
\vspace{-10pt}
\label{tab:full_text}
\setlength{\tabcolsep}{10pt} \scriptsize
\resizebox{\linewidth}{!}{%
\begin{tabular}{lccc}
\toprule
Method & World.$\uparrow$ & Daily.$\uparrow$ & $N_{add}$ \\
\midrule
\rowcolor{gray!8} 
Full Token & 47.7 & 57.7 & -  \\
\quad + text $(t,h,w)$ & 47.9\,\textcolor{ForestGreen}{\scriptsize$(+0.2)$} & 57.3\,\textcolor{red}{\scriptsize$(-0.4)$} & 70 \\
\quad + text $(t,t,t)$ & 47.6\,\textcolor{red}{\scriptsize$(-0.1)$} & 57.8\,\textcolor{ForestGreen}{\scriptsize$(+0.1)$} & 70 \\
\midrule
\rowcolor{gray!8} 
Random & 44.1 & 53.1 & -  \\
\quad + text $(t,h,w)$ & 46.6\,\textcolor{ForestGreen}{\scriptsize$(+2.5)$} & 56.7\,\textcolor{ForestGreen}{\scriptsize$(+3.6)$} & 70 \\
\quad + text $(t,t,t)$ & 46.4\,\textcolor{ForestGreen}{\scriptsize$(+2.3)$} & 56.7\,\textcolor{ForestGreen}{\scriptsize$(+3.6)$} & 70 \\
\bottomrule
\end{tabular}%
}
\vspace{-15pt}
\end{table}
\subpara{Text proxies help only when information is lost.}
\cref{tab:full_text} tests when the text proxies actually help. Added on top of
the full video set, they leave accuracy essentially unchanged, since the
information they carry is already present, so the proxies neither help nor hurt.
The picture changes once tokens are pruned. Added on top of a random $50\%$ prune, the same proxies recover much of the lost accuracy ($+2.5$ and $+3.6$ on World.\ and Daily.), because they reintroduce semantics the pruning had discarded. 
Text proxies are thus useful precisely when information has been removed, not as a generic add-on. Under pruning, placing the proxies at their spatial position $(t,h,w)$ works better than $(t,t,t)$.


\begin{table}[t]
\centering
\caption{\textbf{Text proxy vs.\ revived video tokens.} Under a matched token
budget, expressing detected objects as text proxies outperforms reviving their
video tokens, whether chosen at random or by saliency.}
\vspace{-10pt}
\label{tab:revive}
\setlength{\tabcolsep}{10pt} \footnotesize
\resizebox{\linewidth}{!}{%
\begin{tabular}{lcccc}
\toprule
Method & World.$\uparrow$ & Daily.$\uparrow$ & $N_{add}$ & Comp.$\uparrow$ \\
\midrule
\rowcolor{gray!8} Full Token & 47.7 & 57.7 & - & 0 \\
Random   & 47.2 & 57.2 & 70 & 54 \\
Saliency & 47.3 & 57.3 & 70 & 54 \\
\rowcolor{aliceblue}
\textbf{Ours} & \textbf{47.8} & \textbf{58.3} & 70 & 54 \\
\bottomrule
\end{tabular}%
}
\vspace{-5pt}
\end{table}

\subpara{Why text proxies rather than video tokens.}
A natural alternative to a text proxy is to keep the underlying video tokens
instead. We test this in \cref{tab:revive}. Rather than adding text proxies, we
revive the same number of discarded video tokens ($N_{add}$, the average number
of text tokens ReMo adds) from within detected regions, choosing them
either at random or by our saliency score (\cref{eq:saliency}). This matches the
detected regions and the token budget across all variants, so the only
difference is whether the object is expressed as a few words or as its original
video tokens. Both revival rules fall below ReMo on both benchmarks, reaching
$47.2$ and $47.3$ on World. and $57.2$ and $57.3$ on Daily., whereas ReMo reaches $47.8$ and $58.3$. Naming an object in a few words is thus not only
cheaper than storing its pixels but also more effective, since the proxy states
the semantics directly while the revived tokens still carry redundant appearance detail.

\begin{figure*}[t]
\centering
\includegraphics[width=\textwidth]{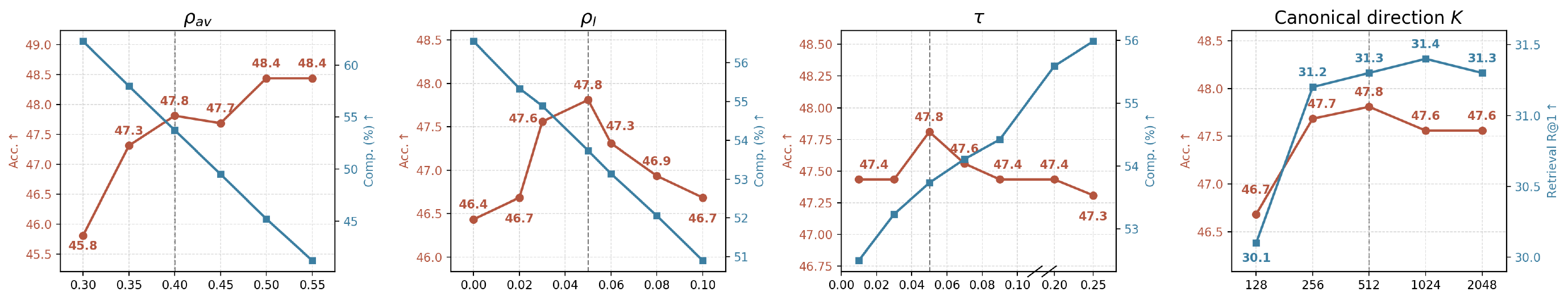}
\vspace{-25pt}
\caption{\textbf{Hyperparameter sensitivity of ReMo}, measured on World.\ with the 3B backbone. We vary the selection budget $\rho_{av}$, local detail budget $\rho_l$, temporal merging threshold $\tau$, and number of canonical directions $K$ individually. Red curves show accuracy and blue curves the compression rate, except retrieval R@1 for $K$. Dashed lines mark the default setting. Accuracy remains stable, showing ReMo is robust to these choices.}
\label{fig:sweep}
\vspace{-15pt}
\end{figure*}

\begin{table}[t]
\centering
\caption{\textbf{Effect of text proxy access.}
We vary which tokens can attend to text proxies on the 3B model:
all tokens, audio-visual tokens only, or text tokens only.
No proxies removes text proxies entirely.
Parenthesized values indicate score changes relative to Ours.}
\vspace{-0.48em}
\label{tab:attention_ablation}
\setlength{\tabcolsep}{7pt}
\scriptsize
\resizebox{\linewidth}{!}{%
\begin{tabular}{lccc}
\toprule
Proxy access & World.$\uparrow$ & Daily.$\uparrow$ & Comp.$\uparrow$(\%) \\
\midrule
\rowcolor{aliceblue}
\textbf{All tokens (Ours)} & \textbf{47.8} & \textbf{58.3} & 54 \\
\midrule
Audio-visual
& 47.2\,\textcolor{red}{\scriptsize$(-0.6)$}
& 57.3\,\textcolor{red}{\scriptsize$(-1.0)$}
& 54 \\
Text
& 47.1\,\textcolor{red}{\scriptsize$(-0.7)$}
& 57.9\,\textcolor{red}{\scriptsize$(-0.4)$}
& 54 \\
No proxies
& 46.7\,\textcolor{red}{\scriptsize$(-1.1)$}
& 57.1\,\textcolor{red}{\scriptsize$(-1.2)$}
& 55 \\
\bottomrule
\end{tabular}%
}
\vspace{-15pt}
\end{table}
\subpara{Who uses the text proxies?}
\cref{tab:attention_ablation} examines which token types benefit from access to the injected text proxies. Allowing all tokens to attend to the proxies performs best on both benchmarks. Restricting access to only audio-visual tokens reduces performance by $0.6$/$1.0$ points on World./Daily., while allowing only text tokens causes drops of $0.7$/$0.4$. Removing the proxies entirely leads to a larger overall degradation despite slightly higher compression. These results suggest that the injected semantics are not used by a single token type, but serve as shared context that benefits both modality and text representations. Additional analysis of attention to the text proxies is provided in~\cref{app:text_attn} of supp. mat.

\begin{table}[t]
\centering
\caption{\textbf{Ablation on object candidate sources.}
$\mathcal{C}_{v}$ and $\mathcal{C}_{a}$ denote object candidates
extracted from visual and audio captions, respectively,
and $\mathcal{Y}$ denotes objects from the detector.}
\vspace{-0.6em}
\label{tab:object_source}
\setlength{\tabcolsep}{9pt}
\scriptsize
\resizebox{\linewidth}{!}{%
\begin{tabular}{lccc}
\toprule
Method & World.$\uparrow$ & Daily.$\uparrow$ & Comp.$\uparrow$(\%) \\
\midrule
\rowcolor{aliceblue}
\textbf{Ours} & 47.8 & \textbf{58.3} & 54 \\
\midrule
All caption
& 43.8\,\textcolor{red}{\scriptsize$(-4.0)$}
& 54.5\,\textcolor{red}{\scriptsize$(-3.8)$}
& 43 \\
$\mathcal{C}_{v} \cap \mathcal{Y}$
& \underline{47.9}\,\textcolor{ForestGreen}{\scriptsize$(+0.1)$}
& \underline{58.1}\,\textcolor{red}{\scriptsize$(-0.2)$}
& 54 \\
$\mathcal{C}_{v} \setminus \mathcal{Y}$
& 47.1\,\textcolor{red}{\scriptsize$(-0.7)$}
& 57.6\,\textcolor{red}{\scriptsize$(-0.7)$}
& 54 \\
$\mathcal{C}_{a} \cap \mathcal{Y}$
& \textbf{48.1}\,\textcolor{ForestGreen}{\scriptsize$(+0.3)$}
& 57.8\,\textcolor{red}{\scriptsize$(-0.5)$}
& 54 \\
$\mathcal{C}_{a} \setminus \mathcal{Y}$
& 46.9\,\textcolor{red}{\scriptsize$(-0.9)$}
& \underline{58.1}\,\textcolor{red}{\scriptsize$(-0.2)$}
& 54 \\
\bottomrule
\end{tabular}%
}
\vspace{-15pt}
\end{table}
\subpara{Object sources for text proxies.}
\cref{tab:object_source} compares full-caption and object-level semantics. We denote objects from visual and audio captions as $\mathcal{C}_v$ and $\mathcal{C}_a$, and detector outputs as $\mathcal{Y}$; intersections retain detected objects, while set differences exclude them. Full-caption injection performs worse, likely because long caption sequences disrupt the interleaved audio-visual structure seen during training. Excluding objects from our detector consistently reduces accuracy on both benchmarks, whereas retaining their intersection yields comparable performance. Candidates from Qwen3.8-27B~\cite{qwen38} and Audio-Flamingo-Next-Instruct~\cite{ghosh2026audio} perform comparably to our lightweight detector but require much larger auxiliary models, favoring the detector's balance of accuracy and efficiency. See~\cref{app:object_source} of supp. mat. for details.

\subpara{Hyperparameter sensitivity.}
\cref{fig:sweep} varies each hyperparameter in turn. The selection budget
$\rho_{av}$ trades accuracy for compression as expected: keeping more video
tokens raises accuracy but lowers the compression rate, so we set
$\rho_{av}{=}0.4$ to match the compression of the training-free baselines rather
than to maximize accuracy. The local detail budget $\rho_l$ peaks around $0.05$,
where restoring a few high-detail tokens helps before extra tokens start to add
redundancy. The temporal-merging threshold $\tau$ is broadly flat, with a mild
peak near $0.05$, showing that ReMo is not sensitive to its exact value. For the number of canonical directions $K$, both retrieval and accuracy saturate by $256$ and stay flat beyond it.
Overall, performance stays within a narrow band across these settings, with mild peaks, which our defaults are set to.

\subpara{Performance under extreme compression.}
\begin{figure}[t]
\centering
\includegraphics[width=\linewidth]{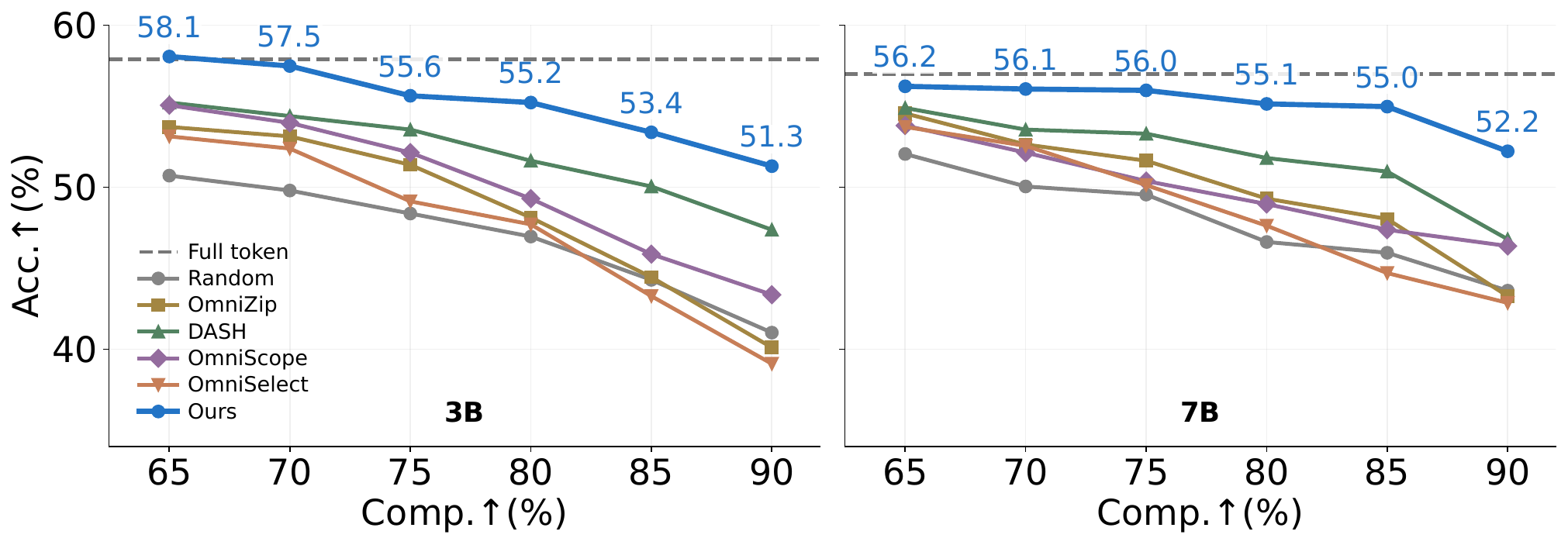}
\vspace{-2.3em}
\caption{\textbf{Extreme compression ratios.} Accuracy under increasingly aggressive token compression on the 3B and 7B backbones, measured on Daily. ReMo degrades more gracefully than competing methods and maintains the strongest performance from $65\%$ to $90\%$ compression.}
\label{fig:comp_ratio}
\vspace{-15pt}
\end{figure}
\cref{fig:comp_ratio} evaluates robustness under increasingly aggressive compression. ReMo degrades gracefully as the compression rate increases and remains above all competing methods across both model scales. At the most aggressive ratios, OmniZip and OmniSelect even fall below random pruning, showing that they struggle under extreme compression. In contrast, at $90\%$ compression, ReMo still achieves $51.3\%$ accuracy on 3B and $52.2\%$ on 7B, substantially narrowing the gap to the full-token model. This shows that ReMo remains effective even under extreme compression. Hyperparameter settings for each baseline and compression ratio are provided in~\cref{app:baseline_settings} of supp. mat.

\section{Conclusion}
We presented \textbf{\emph{ReMo}}, a training-free token compression framework for Omni-LLMs. ReMo removes visual tokens whose information
is already captured by audio through a common-space projection and preserves
object-level semantics through compact text proxies with spatially grounded
positions. On Qwen2.5-Omni at two scales and Qwen3-Omni-30B, ReMo compresses over
half of the input tokens while performing comparably to the full-token model and reducing
memory and latency. ReMo also remains effective on long-video and streaming
benchmarks and degrades gracefully under extreme compression. These results
suggest that cross-modal redistribution preserves multimodal information
with fewer tokens, enabling efficient processing of longer, richer inputs.
Despite our improvements, compression needs vary across tasks and inputs,
making adaptive compression a natural direction for future work.

\section*{LLM Usage Statement}
We used Claude to improve the clarity and grammar of the manuscript and to assist with code refinement. The authors reviewed and revised the suggested edits, verified the resulting code, and take full responsibility for the final manuscript and implementation.

{
    \small
    \bibliographystyle{ieeenat_fullname}
    \bibliography{shortstrings,main}
}

  \savesuppcitationunit
\clearpage
\ifsubmission
\setcounter{page}{1}
\fi

\appendix

\renewcommand{\theHsection}{appendix.\Alph{section}}
\renewcommand{\theHsubsection}{appendix.\Alph{section}.\arabic{subsection}}

\counterwithin{figure}{section}
\counterwithin{table}{section}

\renewcommand{\thefigure}{\thesection.\arabic{figure}}
\renewcommand{\thetable}{\thesection.\arabic{table}}

\makeatletter
\providecommand{\authcount}[1]{}
\makeatother

\twocolumn[
\begin{center}
    \Large \textbf{Out of Sight, Still in Mind: Token Compression for Omni-LLMs} \\
    
    
    \vspace{0.5em}
    \large \textbf{Supplementary Material} \\ 
    \vspace{1.5em} 
\end{center}
]

\startcontents[supp]
\hypersetup{linkcolor=black}
\printcontents[supp]{l}{1}{
\setcounter{tocdepth}{2}
\setlength{\baselineskip}{1\baselineskip}
}
\hypersetup{linkcolor=cvprblue}


\restoresuppcitationunit
\section{ReMo Algorithm}
\label{app:algorithm}
\Cref{alg:remo} summarizes the full ReMo pipeline for a single audio-visual
chunk. The cross-modal projection $(P_a, P_v)$ is built once in closed form
(\cref{sec:proj}) and reused for every chunk. At inference, each chunk is scored
for visual redundancy, pruned to the selection budget, augmented with local
detail and text proxies, and finally merged across adjacent chunks. All steps
are training-free and run before the decoder.

\begin{algorithm}[ht]
\caption{ReMo inference for one chunk $t$}
\label{alg:remo}
\begin{algorithmic}[1]
\Require visual tokens $\{v_{t,j}\}_{j=1}^{N}$, audio token $a_t$, frozen
projections $P_a, P_v$, budgets $\rho_{av}, \rho_l$, threshold $\tau$
\State $z^a_t \gets (a_t-\mu_a)P_a$ \Comment{project audio to common space}
\For{each visual token $v_{t,j}$}
  \State $z^v_{t,j} \gets (v_{t,j}-\mu_v)P_v$
  \State $s^{vis}_{t,j} \gets 1-\cos(v_{t,j},\bar v_t)$ \Comment{intra-visual saliency}
  \State $s^{aud}_{t,j} \gets 1-\cos(z^v_{t,j},z^a_t)$ \Comment{audio saliency}
  \State $s_{t,j} \gets (1-\beta_t)\,s^{vis}_{t,j} + \beta_t\,s^{aud}_{t,j}$
  \Comment{$\beta_t=\bar s^{vis}_t/(\bar s^{vis}_t+\bar s^{aud}_t)$}
\EndFor
\State keep top $\lceil \rho_{av} N \rceil$ tokens by $s_{t,j}$
\State restore top $\lceil \rho_l N \rceil$ discarded tokens by local detail $d_{t,j}$
\State detect object classes; mean-pool each object's subword embeddings into a text proxy and inject it at its mask centroid
\For{co-located tokens $v_{t,j}, v_{t+1,j}$ in adjacent chunks}
  \State $m_j \gets \min(s_{t,j}, s_{t+1,j})$
  \If{$1-\cos(v_{t,j},v_{t+1,j}) < \tau$ \textbf{and} $m_j <$ median of $\{m_j\}$}
    \State $\tilde v_j \gets \dfrac{s_{t,j}\,v_{t,j} + s_{t+1,j}\,v_{t+1,j}}{s_{t,j} + s_{t+1,j}}$
    \Comment{saliency-weighted fusion}
    \State write $\tilde v_j$ to the higher-saliency slot
  \EndIf
\EndFor
\State \Return retained visual, restored, and text-proxy tokens
\end{algorithmic}
\end{algorithm}

\section{Baseline Settings}
\label{app:baseline_settings}
To ensure reproducibility and fair comparison, we restrict baselines to peer-reviewed methods with official code releases.

\subsection{Hyperparameter Settings}
\begin{table}[t]
\centering
\caption{\textbf{Baseline compression settings}, reported as token removal
percentages. Main-evaluation settings use separate audio and visual
budgets in~\cref{tab:main}.
Extreme-compression settings apply equally to both modalities and
both model scales; column headings indicate target overall
compression rates. Dashes indicate settings not evaluated.}
\label{tab:baseline_compression_settings}
\resizebox{\columnwidth}{!}{%
\begin{tabular}{lrrrrrrrr}
\toprule
& \multicolumn{2}{c}{Main evaluation}
& \multicolumn{6}{c}{Extreme compression: Comp. (\%)} \\
\cmidrule(lr){2-3}
\cmidrule(lr){4-9}
Method & Audio & Visual & 65 & 70 & 75 & 80 & 85 & 90 \\
\midrule
FastV~\cite{chen2024image}
& 50 & 50
& -- & -- & -- & -- & -- & -- \\
OmniZip~\cite{tao2026omnizip}
& 30.0 & 60.0
& 68.5 & 73.5 & 79.0 & 84.0 & 89.0 & 94.5 \\
DASH~\cite{li2026dash}
& 45.0 & 47.5
& 59.0 & 64.5 & 70.5 & 76.0 & 82.0 & 87.5 \\
OmniScope~\cite{su2026omniscope}
& 25.0 & 57.5
& 65.0 & 70.5 & 76.0 & 81.5 & 86.5 & 91.5 \\
OmniSelect~\cite{yang2026omniselect}
& 55.0 & 55.0
& 66.0 & 71.0 & 76.0 & 81.0 & 86.0 & 91.0 \\
\bottomrule
\end{tabular}%
}
\end{table}
\cref{tab:baseline_compression_settings} summarizes the compression
settings used in the main~\cref{tab:main} and extreme compression evaluations~\cref{fig:comp_ratio}.
For consistency across implementations, we report all settings as
token removal percentages rather than retention percentages,
converting retention-based parameters where necessary.
These values specify compression budgets and should be distinguished
from the measured overall input token compression rates.
For the main evaluation, we use separate audio and visual budgets
following the default configurations of the respective repositories
listed in \cref{tab:baseline_repositories}.
Since the repositories do not specify settings for our target extreme
compression rates, we adjust each method's removal percentage to match
these targets, applying the same percentage to audio and visual tokens
and using identical settings for the 3B and 7B backbones.
For ReMo, we vary only $\rho_{av}$, setting it to
$0.295$, $0.235$, $0.175$, $0.115$, $0.045$, and $0.010$
for target compression rates of 65\%, 70\%, 75\%, 80\%, 85\%,
and 90\%, respectively, while fixing $\rho_l=0.05$ and $\tau=0.25$ (raised from the
default 0.05 to allow more temporal merging at high compression).

\begin{table}[t]
\centering
\caption{\textbf{Baseline repositories and compression controls.}
Audio and visual parameter names are listed in that order.}
\label{tab:baseline_repositories}
\resizebox{\columnwidth}{!}{%
\begin{tabular}{@{}lp{0.29\textwidth}p{0.50\textwidth}@{}}
\toprule
Method & Repository & Compression controls \\
\midrule
FastV~\cite{chen2024image}
& \href{https://github.com/pkunlp-icler/FastV}
       {\texttt{pkunlp-icler/FastV}}
& \texttt{fastv\_r} \\

OmniZip~\cite{tao2026omnizip}
& \href{https://github.com/KD-TAO/OmniZip}
       {\texttt{KD-TAO/OmniZip}}
& \texttt{merging\_ratio\_audio},
  \texttt{merging\_ratio\_v} \\

DASH~\cite{li2026dash}
& \href{https://github.com/laychou666/DASH}
       {\texttt{laychou666/DASH}}
& \texttt{merging\_ratio\_audio},
  \texttt{merging\_ratio\_v} \\

OmniScope~\cite{su2026omniscope}
& \href{https://github.com/MAC-AutoML/OmniScope}
       {\texttt{MAC-AutoML/OmniScope}}
& \texttt{audio\_prune\_ratio},
  \texttt{visual\_prune\_ratio} \\

OmniSelect~\cite{yang2026omniselect}
& \href{https://github.com/Yangmrl-nlp/OmniSelect}
       {\texttt{Yangmrl-nlp/OmniSelect}}
& \texttt{prune\_ratio\_a},
  \texttt{prune\_ratio\_v} \\
\bottomrule
\end{tabular}%
}
\end{table}
\subsection{Implementation and Parameter Mapping}
\cref{tab:baseline_repositories} lists the repositories used for
the baselines and the corresponding compression controls.
For OmniZip and DASH, the audio and visual budgets correspond to
\texttt{merging\_ratio\_audio} and \texttt{merging\_ratio\_v};
for OmniScope, to \texttt{audio\_prune\_ratio} and
\texttt{visual\_prune\_ratio}; and for OmniSelect, to
\texttt{prune\_ratio\_a} and \texttt{prune\_ratio\_v}.
All percentages in \cref{tab:baseline_compression_settings}
are reported as token removal rates.
All other settings follow the defaults in the respective repositories in~\cref{tab:baseline_repositories}.

\begin{table}[t]
\centering
\caption{\textbf{External module sizes} for Qwen2.5-Omni$_{7\mathrm{B}}$.
Model parameters and CCA matrix entries are reported separately,
with their sum shown in the Total column. Relative size is normalized to Ours.}
\label{tab:external_modules}
\resizebox{\columnwidth}{!}{%
\begin{tabular}{llrrrr}
\toprule
Method & External modules &  Params. & CCA Mat. & Total & Rel. \\
\midrule
\rowcolor{aliceblue}
\textbf{Ours}
& YOLOv8n-seg~\cite{yaseen2024yolov8indepthexplorationinternal} & 3.4M & 3.7M & \textbf{7.1M} & 1.0$\times$ \\
\midrule
OmniScope~\cite{su2026omniscope}
& CLIP ViT-L/14@336~\cite{radford2021learning} & 427.9M & - & 427.9M & 60.3$\times$ \\
OmniSelect~\cite{yang2026omniselect}
& AudioCLIP~\cite{guzhov2022audioclip} & 134.1M & - & 134.1M & 18.9$\times$ \\
\bottomrule
\end{tabular}%
}
\end{table}
\subsection{External Module Sizes}
\label{app:external_module}
\cref{tab:external_modules} compares the sizes of the external modules.
ReMo uses a lightweight YOLOv8n detector and CCA projections,
totaling 5.5M and 7.1M parameters and coefficients for the 3B and
7B backbones, respectively. This is substantially smaller than
OmniScope's CLIP encoder (427.9M) and OmniSelect's AudioCLIP
encoder (134.1M), limiting the additional resource cost.
Consistent with this smaller overhead, ReMo achieves lower memory
usage and latency than both methods in~\cref{tab:efficiency}.

\section{Benchmark Statistics}
\begin{table}[t]
\centering
\caption{\textbf{Average input token counts by modality.}
Percentages in parentheses indicate each modality's share of the total.
The overall average is weighted by the number of questions.}
\label{tab:input_token_statistics}
\resizebox{\columnwidth}{!}{%
\begin{tabular}{lrrrrr}
\toprule
Benchmark & Questions & Total & Visual & Audio & Text \\
\midrule
WorldSense~\cite{hong2025worldsense}
& 799 & 14,126.3
& 12,842.5 (90.9\%) & 1,106.0 (7.8\%) & 177.8 (1.3\%) \\
DailyOmni~\cite{zhou2025daily}
& 1,197 & 13,787.0
& 12,520.4 (90.8\%) & 1,080.7 (7.8\%) & 185.9 (1.3\%) \\
Video-MME~\cite{fu2025video}
& 231 & 30,172.7
& 28,898.4 (95.8\%) & 1,107.5 (3.7\%) & 166.7 (0.6\%) \\
OmniVideoBench~\cite{li2025omnivideobench}
& 166 & 24,162.8
& 23,161.3 (95.9\%) & 815.1 (3.4\%) & 186.4 (0.8\%) \\
\midrule
\textbf{Overall average}
& \textbf{2,393} & \textbf{16,201.8}
& \textbf{14,947.1 (92.3\%)}
& \textbf{1,073.3 (6.6\%)}
& \textbf{181.4 (1.1\%)} \\
\bottomrule
\end{tabular}%
}
\end{table}
\subsection{Input Token Statistics}
\cref{tab:input_token_statistics} shows that visual tokens account for
over 90\% of the input tokens on every benchmark and 92.3\% overall.
This imbalance motivates targeting visual tokens for compression,
where reducing redundancy can yield the largest token savings.

\begin{table}[t]
\centering
\caption{\textbf{Video duration statistics for the evaluated benchmarks.}
All durations are reported in seconds.}
\label{tab:video_duration_statistics}
\resizebox{\columnwidth}{!}{%
\begin{tabular}{lrrrrrrr}
\toprule
Benchmark & Videos & Samples & Mean & Median & Std. & Min. & Max. \\
\midrule
WorldSense~\cite{hong2025worldsense}
& 403 & 799 & 44.4 & 44.8 & 10.2 & 15.6 & 60.0 \\
DailyOmni~\cite{zhou2025daily}
& 684 & 1,197 & 42.8 & 30.0 & 15.5 & 9.0 & 60.0 \\
Video-MME~\cite{fu2025video}
& 77 & 231 & 44.3 & 47.0 & 12.9 & 11.0 & 60.0 \\
OmniVideoBench~\cite{li2025omnivideobench}
& 96 & 166 & 31.2 & 30.4 & 14.6 & 4.9 & 59.8 \\
\midrule
\textbf{Overall}
& \textbf{1,260} & \textbf{2,393}
& \textbf{42.5} & \textbf{41.6} & \textbf{14.2}
& \textbf{4.9} & \textbf{60.0} \\
\bottomrule
\end{tabular}%
}
\end{table}
\subsection{Video Duration Statistics}
\cref{tab:video_duration_statistics} summarizes the durations of the
evaluated videos. The evaluation includes 1,260 videos and 2,393 samples,
with durations ranging from 4.9 to 60.0 seconds and an overall mean of
42.5 seconds. We limit video duration to at most one minute due to
memory constraints when running full-token inference on a single GPU.

\section{Long-Video and Streaming Evaluation}
\label{app:long_streaming}
For long-video evaluation, we use the long-video subset of
Video-MME~\cite{fu2025video} and LVOmniBench~\cite{tao2026lvomnibench}.
For streaming evaluation, we use the Omni subset of
StreamingBench~\cite{lin2026streamingbench},
FutureOmni~\cite{chen2026futureomni}, and the SG, AP, MD, and SI
tasks of OmniMMI~\cite{wang2025omnimmi}.
We evaluate only ReMo in the streaming setting, as existing
methods require information unavailable during online processing.
OmniZip and DASH compute audio saliency over the entire input
to select globally important audio tokens, which then guide
visual token selection.
In contrast, FastV, OmniScope, and OmniSelect perform
query-dependent pruning, which cannot be applied before
the query arrives.
ReMo instead supports online pruning without requiring
future audio or advance knowledge of the query.
Duration statistics and audio-visual preprocessing details
are provided below.

\begin{table}[t]
\centering
\caption{\textbf{Duration statistics for long-video and streaming evaluation.}
Long-video statistics describe source video durations, while streaming
statistics describe the actual input segment for each question.
All durations are reported in minutes.}
\label{tab:long_streaming_statistics}
\resizebox{\columnwidth}{!}{%
\begin{tabular}{lrrrrrrr}
\toprule
Benchmark & Videos & Samples & Mean & Median & Std. & Min. & Max. \\
\midrule
\multicolumn{8}{c}{\textit{Long-video}} \\
\midrule
Video-MME Long
& 300 & 900 & 41.1 & 40.4 & 8.1 & 29.9 & 59.7 \\
LVOmniBench~\cite{tao2026lvomnibench}
& 275 & 1,014 & 34.6 & 32.1 & 15.2 & 10.2 & 91.4 \\
\midrule
\multicolumn{8}{c}{\textit{Streaming}} \\
\midrule
StreamingBench~\cite{lin2026streamingbench}
& 200 & 1,000 & 3.1 & 2.6 & 2.2 & 0.1 & 14.8 \\
FutureOmni~\cite{chen2026futureomni}
& 1,034 & 1,034 & 2.7 & 2.3 & 2.0 & 0.02 & 17.8 \\
OmniMMI~\cite{wang2025omnimmi}
& 712 & 1,809 & 2.4 & 0.3 & 7.3 & 0.007 & 98.6 \\
\bottomrule
\end{tabular}%
}
\end{table}
\subsection{Video Duration Statistics}
\cref{tab:long_streaming_statistics} summarizes the durations used
in our long-video and streaming evaluations.
The long-video benchmarks contain videos averaging 41.1 and
34.6 minutes, with a maximum duration of 91.4 minutes.
For streaming benchmarks, we report the actual input segment
available for each question rather than the full source video.
These segments vary substantially in length, ranging from
0.007 to 98.6 minutes across benchmarks.

\subsection{Audio-Visual Preprocessing}
The Qwen2.5-Omni~\cite{xu2025qwen25omnitechnicalreport} preprocessing implementation used in our setup
truncates audio to five minutes\footnote{The official
\href{https://huggingface.co/Qwen/Qwen2.5-Omni-7B/blob/main/preprocessor_config.json}
{preprocessing configuration}
sets \texttt{chunk\_length} to 300 seconds.}.
Video frames are sampled at 2 FPS by default, with a maximum of
768 frames\footnote{See the official
\href{https://github.com/QwenLM/Qwen2.5-Omni/blob/main/qwen-omni-utils/src/qwen_omni_utils/v2_5/vision_process.py}
{video preprocessing implementation}.}.
When this limit is exceeded, frames are sampled uniformly across
the input span at a lower effective frame rate.
To maintain local audio-visual correspondence, we extract a
0.5-second audio segment at each sampled frame's timestamp and
concatenate these segments in chronological order.
We then process the sampled frames and concatenated audio as a
2-FPS sequence, preserving the frame-to-audio correspondence
of the standard input format while covering the full input span.
All experiments are run on four NVIDIA RTX A6000 GPUs to avoid
out-of-memory errors during full token inference.

\section{Further Analysis}
\label{app:text}


\subsection{Effect of the Projection on Selection}
Before projection, raw audio saliency exhibits a correlation with visual
saliency ($\rho{=}0.32$ on Daily. and $0.50$ on MME). After projection, this
correlation drops to nearly zero ($0.02$ and $0.00$), while projected audio
saliency becomes almost uncorrelated with raw audio saliency
($\rho{=}0.01$ and $-0.01$). 
The projection therefore provides a complementary
cross-modal signal rather than simply refining the raw audio ranking. As shown
in \cref{tab:salmode}, neither visual nor audio saliency alone is sufficient,
whereas combining them performs best. Replacing audio saliency with random
scores also reduces performance. Together with the degradation without
projection in \cref{tab:ablation}, this suggests that the projection helps
audio saliency effectively complement the visual signal.
Accordingly, the residual-saliency score
in \cref{eq:saliency} fuses projected-audio saliency with visual saliency,
using the former as a complementary cue.

\begin{table}[t]
\centering
\caption{\textbf{Saliency ranking source} on the 3B model.
We compare vision-only ($\beta{=}0$), raw-audio only, and
projected-audio only ($\beta{=}1$) ranking.
Random audio saliency replaces $s^{\mathrm{aud}}$ with samples
from $\mathcal{U}(0,1)$ while retaining the visual saliency term.
Combining visual and projected audio saliency performs best.}
\label{tab:salmode}
\setlength{\tabcolsep}{20pt}\small
\resizebox{\linewidth}{!}{%
\begin{tabular}{lcc}
\toprule
Saliency source & Daily.$\uparrow$ & MME$\uparrow$ \\
\midrule
\rowcolor{gray!8} Full Token & 57.7 & 75.8 \\
Vision only ($\beta{=}0$)      & 57.8 & 75.8 \\
Raw-audio only ($\beta{=}1$)   & 56.0 & 74.5 \\
Proj.-audio only ($\beta{=}1$)  & 56.1 & 76.6 \\
Random audio sal. & 57.4 & 75.8 \\
\rowcolor{aliceblue}
\textbf{Ours}          & \textbf{58.3} & \textbf{77.5} \\
\bottomrule
\end{tabular}%
}
\end{table}
\begin{table}[t]
\centering
\vspace{5pt}
\caption{\textbf{Raw vs.\ canonical space (Ours) video token selection.} Under the
same budget, we compare the tokens kept by raw audio-video similarity and by our
common-space alignment on 3B model. Overlap is the Jaccard similarity of the two kept sets,
and Flipped is the fraction of tokens whose keep/drop decision changes. The
projection preserves most of the selection and flips only a thin margin,
consistently across benchmarks.}
\label{tab:ccaw_overlap}
\setlength{\tabcolsep}{20pt}\small
\resizebox{\linewidth}{!}{%
\begin{tabular}{lcc}
\toprule
Benchmark & Overlap (Jaccard) & Flipped \\
\midrule
MME        & 83.1 & 6.9\% \\
World.      & 83.7 & 6.7\% \\
OmniVid.    & 83.3 & 6.9\% \\
Daily.      & 83.2 & 6.8\% \\
\midrule
Mean & 83.3 & 6.8\% \\
\bottomrule
\end{tabular}
}
\end{table}

Under the selection induced by \cref{eq:saliency}, \cref{tab:ccaw_overlap}
examines how replacing raw audio-video similarity with common-space similarity
changes the selected tokens. At the same budget, the two selections share a
Jaccard overlap of $83.3\%$, with only $6.8\%$ of keep/drop decisions differing.
These values remain stable across benchmarks, varying by at most $0.6$ points
in overlap and $0.2\%$ in flipped decisions. Projection thus preserves clear-cut
decisions and changes only a narrow set of tokens near the selection boundary.
This explains why its substantial retrieval improvement translates into a more
modest downstream accuracy gain: in the fused residual-saliency score, improved
audio-video alignment affects only a small fraction of decisions, yielding the
consistent gains in \cref{tab:ablation}.

\begin{figure*}[t]
    \centering
    \includegraphics[width=\textwidth]{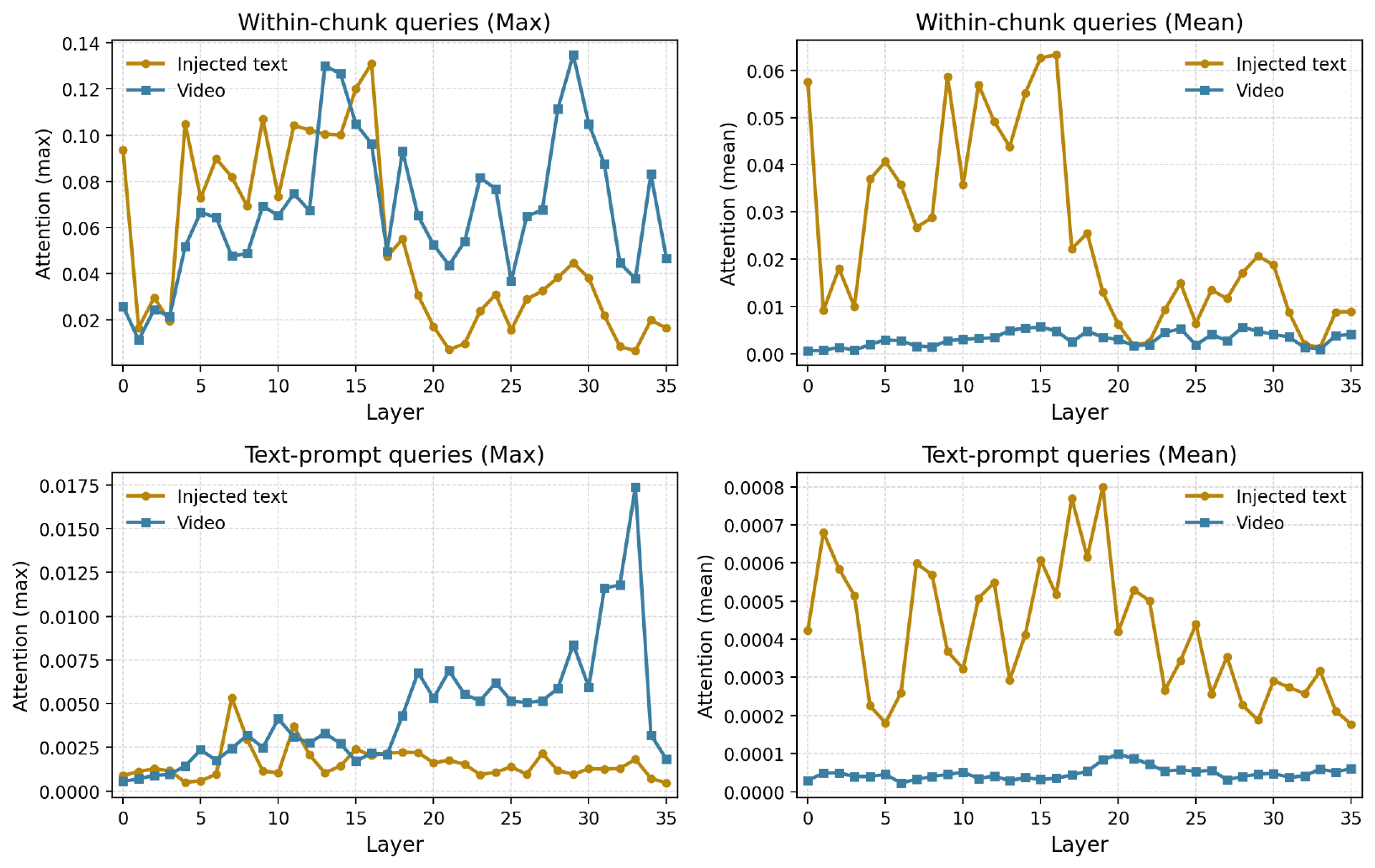}
    \vspace{-25pt}
    \caption{\textbf{Attention on the injected text proxies} across decoder layers on
the 3B model. In mean attention, both within-chunk and text-prompt queries
attend more to the injected text than to the video tokens across most layers.
The maximum attention varies by layer, but the consistently higher mean indicates
that the proxies are broadly attended, not reached only by a few outlier heads.}
\label{fig:text_attn}
\vspace{-15pt}
\end{figure*}

\subsection{Token Budget}
\cref{tab:token_budget} reports how many tokens the text proxies actually cost.
Averaged over the four benchmarks, a sample carries about $19{,}355$ video tokens, of which ReMo drops $59\%$. The dropped semantics are reintroduced with only $65$ class-text tokens on average, one per detected class. The ratio of dropped video tokens to injected text tokens is roughly $176{:}1$, meaning that each injected text token summarizes the shared semantics of a large set of redundant visual tokens rather than replacing them one by one. This is the concrete sense in which a few words replace many video tokens.
\begin{table}[t]
\centering
\caption{\textbf{Token budget of the text proxy} on the 3B model, averaged
over four benchmarks (World., Daily., MME, and OmniVid.). ReMo drops most video tokens and reintroduces the pruned
semantics with a tiny set of class-text tokens, one per detected objects, about a
$176{:}1$ reduction.}
\label{tab:token_budget}
\setlength{\tabcolsep}{20pt}\small
\resizebox{\linewidth}{!}{%
\begin{tabular}{lc}
\toprule
Quantity (per sample) & Count \\
\midrule
Video tokens (input)        & 19{,}355 \\
\quad$\hookrightarrow$ dropped   & 11{,}425 \; (59.0\%) \\
\quad$\hookrightarrow$ kept      & \phantom{0}7{,}930 \; (41.0\%) \\
\midrule
Text tokens injected (1/object) & 65 \\
\midrule
Dropped video : injected text  & \textbf{176 : 1} \\
\bottomrule
\end{tabular}%
}
\vspace{-15pt}
\end{table}

\subsection{Where the Injected Text is Attended}
\label{app:text_attn}
\cref{fig:text_attn} traces the attention received by the injected text tokens
across decoder layers. In mean attention, the injected text draws well above the
video tokens under within-chunk queries through the early and middle layers, and
again under the final text-prompt queries across most layers, indicating that the
model reads the proxies rather than ignoring them. The maximum attention varies
by layer, sometimes favoring video, but the consistently higher mean shows that
the proxies are broadly attended rather than reached only by a few outlier heads.
The proxies are thus actively used by the model, not inert padding, throughout decoding. This confirms effective semantic preservation.

\begin{table}[t]
\centering
\caption{\textbf{Detector-size ablation.} Accuracy is robust across YOLO backbones;
only the detector's marginal latency cost and its own GPU footprint grow with
size, so ReMo uses the smallest (Nano).}
\label{tab:yolo_eff}
\setlength{\tabcolsep}{10pt}\small
\resizebox{\linewidth}{!}{%
\begin{tabular}{lcccc}
\toprule
Method & \#Param & Mem (GiB)$\downarrow$ & Latency (s)$\downarrow$ & Acc.$\uparrow$ \\
\midrule
\rowcolor{aliceblue}
Nano   & 3.4M  & 0.44 & 0.12 & 77.5 \\
Small  & 11.8M & 0.70 & 0.23 & 78.4 \\
Medium & 27.3M & 1.26 & 0.35 & 78.4 \\
Large  & 46.0M & 1.33 & 0.43 & 77.5 \\
\bottomrule
\end{tabular}
}
\end{table}
\subsection{Detector Size}
\cref{tab:yolo_eff} varies the YOLO backbone from Nano to Large on MME. Accuracy is
essentially flat across the four sizes, moving within one point ($77.5$ to
$78.4$) with no consistent trend, so a larger detector brings no reliable gain.
What does grow with size is cost: memory rises from $0.44$ to $1.33$ GiB and the
detector's latency from $0.12$ to $0.43$ seconds. Since the text proxies depend
only on the detected object classes, which the smallest model already recovers,
ReMo uses YOLOv8n and spends its budget on the LLM rather than the detector.

\begin{table}[t]
\centering
\caption{\textbf{Local-detail connectivity.} Spatial-variance foreground
recovery over 4- vs.\ 8-connected neighbours on 3B model. The 4-neighbour
rule is better.}
\label{tab:abl_conn}
\label{tab:abl_neigh}
\setlength{\tabcolsep}{25pt}\small
\resizebox{\linewidth}{!}{%
\begin{tabular}{lcc}
\toprule
Connectivity & World.$\uparrow$ & Daily.$\uparrow$ \\
\midrule
\rowcolor{aliceblue}\textbf{4-neighbour} & \textbf{47.8} & \textbf{58.3} \\
8-neighbour & 46.6 & 57.9 \\
\bottomrule
\end{tabular}%
}
\end{table}
\subsection{Neighbour Connectivity for Local Detail.}
The local-detail score $d_{t,j}$ (\cref{eq:local}) measures how much a token
differs from its spatial neighbours. \cref{tab:abl_conn} compares using the four
directly adjacent neighbours against the eight surrounding ones. The four-neighbour
version is better on both benchmarks ($47.8$ vs.\ $46.6$ on World.), so ReMo
uses it. The eight-neighbour variant blurs the score by averaging over diagonal
tokens that are less directly related, weakening its ability to flag genuine
local structure.

\begin{table}[t]
\centering
\caption{\textbf{Temporal-fusion destination} on the 3B model. Position to
which the fused token is written across matched frames. Writing to the
highest-saliency slot is best or tied-best.}
\label{tab:abl_dest}
\setlength{\tabcolsep}{25pt}\small
\resizebox{\linewidth}{!}{%
\begin{tabular}{lcc}
\toprule
Destination & World.$\uparrow$ & Daily.$\uparrow$ \\
\midrule
\rowcolor{aliceblue} 
Max saliency & \textbf{47.8} & \textbf{58.3} \\
Fixed $i$ & 47.6 & 58.1 \\
Min saliency & \textbf{47.8} & 57.3 \\
\bottomrule
\end{tabular}%
}
\end{table}
\subsection{Destination of the Temporal Fusion}
When two co-located tokens are merged across adjacent chunks, the fused token
must be written to one position. \cref{tab:abl_dest} compares writing it to the
higher-saliency slot, the earlier chunk's slot regardless of saliency
(Fixed~$i$), or the lower-saliency slot. Writing to the maximum-saliency
slot is best or tied-best, since it keeps the fused content anchored to the more
informative of the two frames. ReMo adopts this choice.

\begin{table}[t]
\centering
\caption{\textbf{Effect of text proxy content.}
Random Gaussian replaces text proxies with Gaussian embeddings
rescaled to match the original proxy norms.
Results are reported on the 3B model.
Parenthesized values indicate score changes relative to Ours.}
\label{tab:proxy_semantics}
\setlength{\tabcolsep}{15pt}
\scriptsize
\resizebox{\linewidth}{!}{%
\begin{tabular}{lcc}
\toprule
Proxy variant & World.$\uparrow$ & Daily.$\uparrow$ \\
\midrule
\rowcolor{aliceblue}
\textbf{Ours}
& \textbf{47.8}
& \textbf{58.3} \\
\midrule
No proxies
& 46.7\,\textcolor{red}{\scriptsize$(-1.1)$}
& 57.1\,\textcolor{red}{\scriptsize$(-1.2)$} \\
Random Gaussian
& 46.4\,\textcolor{red}{\scriptsize$(-1.4)$}
& 57.1\,\textcolor{red}{\scriptsize$(-1.2)$} \\
\bottomrule
\end{tabular}%
}
\end{table}
\subsection{Do Text Proxies Merely Act as Attention Sinks?}
To examine whether text proxies merely serve as attention sinks,
we replace them with random Gaussian embeddings rescaled to
match the norms of the original proxy embeddings.
We compare this variant with ReMo and a variant that removes
text proxies entirely on the 3B model.

As shown in \cref{tab:proxy_semantics}, Gaussian proxies reduce
accuracy by 1.4 points on World. and 1.2 points on Daily.
relative to ReMo, performing comparably to or worse than
removing proxies entirely.
These results suggest that the benefit of text proxies depends
on their semantic content, rather than solely on their presence
as additional attention anchors.

\section{Qwen3-Omni Details}
\label{app:q3}

\subsection{Setup}
We evaluate ReMo on Qwen3-Omni-30B~\cite{xu2025qwen3} (Instruct and Thinking),
running all experiments on a single NVIDIA RTX PRO 6000
Blackwell (96GB) GPU. We use the same hyperparameter settings
as for Qwen2.5-Omni. We report only OmniZip among the competing
methods, as it is the only baseline with publicly available
code supporting Qwen3-Omni.

\begin{table*}[!t]
\centering
\small
\setlength{\tabcolsep}{20pt}
\caption{\textbf{Results on additional backbones.}
Pruned reports the percentage of tokens removed.
Avg. reports the mean benchmark accuracy normalized
to the full-token model.}
\label{tab:additional_backbones}
\resizebox{\linewidth}{!}{%
\begin{tabular}{lrrrrrr}
\toprule
Method & Comp.$\uparrow$(\%) & World.$\uparrow$ & Daily.$\uparrow$
& MME$\uparrow$ & OmniVid.$\uparrow$
& Avg.$\uparrow$(\%) \\
\midrule
\multicolumn{7}{c}{\textit{SALMONN2+$_{7\mathrm{B}}$}} \\
\midrule
\rowcolor{gray!8}
Full Token & 0 & 49.1 & 53.6 & 81.4 & 42.2 & 100.0 \\
Random & 50 & 45.8 & 50.7 & 78.8 & 38.0 & 93.7 \\
\rowcolor{aliceblue}
\textbf{Ours} & 59 & \textbf{47.3} & \textbf{52.2}
& \textbf{80.5} & \textbf{41.0} & \textbf{97.5} \\
\midrule
\multicolumn{7}{c}{\textit{SALMONN2-Pro$_{8\mathrm{B}}$}} \\
\midrule
\rowcolor{gray!8}
Full Token & 0 & 56.8 & 72.8 & 81.4 & 52.4 & 100.0 \\
Random & 50 & 53.6 & 66.2 & 80.5 & 48.8 & 94.3 \\
\rowcolor{aliceblue}
\textbf{Ours} & 58 & \textbf{54.9} & \textbf{71.7}
& \textbf{81.0} & \textbf{50.6} & \textbf{97.8} \\
\bottomrule
\end{tabular}%
}
\end{table*}

\begin{table}[t]
\centering
\caption{\textbf{Qwen3-Omni-30B-A3B-Thinking configuration.} We follow the
official recommended decoding settings for the Thinking variant. The Instruct
variant uses greedy decoding.}
\label{tab:q3_config}
\setlength{\tabcolsep}{25pt}\small
\begin{tabular}{lc}
\toprule
Parameter & Value \\
\midrule
temperature      & 0.6 \\
top-$p$          & 0.95 \\
top-$k$          & 20 \\
max\_tokens   & 1{,}024 \\
seed             & 42 \\
\bottomrule
\end{tabular}%
\end{table}
\begin{table}[t]
\centering
\caption{\textbf{Audio-to-video retrieval on VGGSound} for Qwen3-Omni-30B. For each audio query we rank candidate videos, contrasting raw embeddings
(\emph{Raw}) with our projected ones (\emph{Ours}). The same alignment gain observed at the smaller scale carries over here, so cross-modal redundancy stays reliably measurable in the common space.}
\label{tab:retrieval_qwen3}
\setlength{\tabcolsep}{15pt}\small
\resizebox{\linewidth}{!}{%
\begin{tabular}{lcccc}
\toprule
Method & R@1$\uparrow$ & R@5$\uparrow$ & R@10$\uparrow$ & MedR$\downarrow$ \\
\midrule
\multicolumn{5}{c}{\textit{Qwen3-Omni-30B-A3B (Instruct)}} \\
\midrule
Raw          &  6.3 & 18.5 & 28.4 & 31 \\
\rowcolor{aliceblue}
\textbf{Ours} & \textbf{27.2} & \textbf{54.3} & \textbf{66.0} & \textbf{4} \\
\midrule
\multicolumn{5}{c}{\textit{Qwen3-Omni-30B-A3B (Thinking)}} \\
\midrule
Raw          &  6.1 & 18.7 & 28.3 & 33 \\
\rowcolor{aliceblue}
\textbf{Ours} & \textbf{27.6} & \textbf{54.5} & \textbf{66.0} & \textbf{4} \\
\bottomrule
\end{tabular}%
}
\vspace{-15pt}
\end{table}
\subsection{Configuration}
\cref{tab:q3_config} lists the decoding configuration for the Qwen3-Omni-30B-A3B
Thinking variant, which follows the official recommended sampling settings with
the seed fixed to $42$. The Instruct variant uses greedy decoding. Generation is
capped at $1{,}024$ new tokens, which is ample for the short answers in our
benchmarks. For the Thinking variant, we parse the answer that follows the
reasoning block. If the reasoning does not close within the $1{,}024$-token
budget, we append the bridging cue below to force-close the reasoning block and
let the model generate a brief conclusion for up to $64$ more tokens, so that a
parsable answer is always obtained.

\begin{tcolorbox}[colback=gray!5,colframe=gray!40,boxrule=0.5pt,
  left=6pt,right=6pt,top=4pt,bottom=4pt]
\ttfamily\small
Considering the limited time by the user, I have to give the solution based on
the thinking directly now.\\
$\langle$/think$\rangle$
\end{tcolorbox}

\subsection{Cross-modal Projection}
To check that the projection generalizes beyond Qwen2.5-Omni, we repeat the
audio-to-video retrieval analysis (\cref{sec:proj}) on Qwen3-Omni-30B.
\cref{tab:retrieval_qwen3} shows the same pattern as the smaller backbone. In the
raw embedding space audio and video are nearly unrelated, while after our
projection retrieval improves sharply and the median rank drops to the low
single digits. The common-space alignment that ReMo relies on therefore holds at
the larger scale as well.

\section{Evaluation on Additional Backbones}
\cref{tab:additional_backbones} evaluates ReMo on SALMONN2+ and SALMONN2-Pro~\cite{tang2025video}\footnote{Official implementation: \url{https://github.com/bytedance/video-SALMONN-2}}
across four audio-visual benchmarks, using the same hyperparameters as in our
main experiments..
Since both the publicly available implementations and the evaluations reported in the other baselines' papers are limited to Qwen2.5-Omni, we compare ReMo with random pruning.

In our evaluation, both SALMONN backbones process video
at 10 FPS, compared with 2 FPS for the Qwen backbones,
yielding substantially more visual tokens and higher
compression ratios under the same hyperparameter settings.
ReMo removes 59\% and 58\% of tokens on SALMONN2+ and
SALMONN2-Pro, respectively, while retaining 97.5\% and
97.8\% of the full-token models' average accuracy.
Although ReMo does not surpass full-token performance,
it outperforms random pruning by 3.8 and 3.5
percentage points, respectively, despite removing more
tokens than the 50\% random baseline.
These results support the applicability of cross-modal
redistribution beyond the Qwen backbones.

\section{Caption Generation and Injection}
\label{app:object_source}

We evaluate visual and audio captions as alternatives to the
object text extracted by YOLO~\cite{yaseen2024yolov8indepthexplorationinternal} in ReMo.
We use Qwen3.8-27B~\cite{qwen38} for visual captioning and
Audio-Flamingo-Next-Instruct~\cite{ghosh2026audio}
for audio captioning.

\subsection{Visual Caption Generation}
For each temporal patch containing two frames, we caption
the first frame using the frame indices stored in the YOLO cache.
Qwen3.8-27B receives a single image and the following prompt.

\begin{tcolorbox}[colback=gray!5,colframe=gray!40,boxrule=0.5pt,
  left=6pt,right=6pt,top=4pt,bottom=4pt]
\ttfamily\small
Describe the visible objects and actions in this image
in one or two sentences.
\end{tcolorbox}

We do not provide downstream questions, answer choices,
ground truth answers, or YOLO labels during caption generation.
Frame indices and captions are cached for each video and reused
across questions associated with the same video.

\subsection{Audio Caption Generation}
We generate audio captions using
Audio-Flamingo-Next-Instruct
with the following prompt:

\begin{tcolorbox}[colback=gray!5,colframe=gray!40,boxrule=0.5pt,
  left=6pt,right=6pt,top=4pt,bottom=4pt]
\ttfamily\small
Describe the audible sounds and their sources
in one or two sentences.
\end{tcolorbox}

\subsection{Proxy Injection}
We replace the object text extracted by YOLO with the generated
captions, injecting all caption subword embeddings in their
original order without pooling or truncation.
Reported compression rates include all injected caption tokens (\cref{tab:object_source}).

\subsection{Object Separation}
To separate entities mentioned in captions according to
YOLO detections, we append a follow-up prompt to the same
conversation after caption generation.
For visual captions, Qwen3.8-27B receives:

\begin{tcolorbox}[colback=gray!5,colframe=gray!40,boxrule=0.5pt,
  left=6pt,right=6pt,top=4pt,bottom=4pt]
\ttfamily\small
YOLO detected: \{objects\}\\
List the visible physical objects mentioned in your caption:
"yolo" for matches to YOLO (including synonyms; use YOLO labels),
"caption" for the rest.\\
Return only JSON: \{"yolo": [], "caption": []\}.
\end{tcolorbox}

For audio captions, Audio-Flamingo-Next-Instruct receives:

\begin{tcolorbox}[colback=gray!5,colframe=gray!40,boxrule=0.5pt,
  left=6pt,right=6pt,top=4pt,bottom=4pt]
\ttfamily\small
YOLO detected: \{objects\}\\
List the physical sound sources described in your caption:
"yolo" for matches to YOLO (including synonyms; use YOLO labels),
"caption" for the rest.\\
Return only JSON: \{"yolo": [], "caption": []\}
\end{tcolorbox}

Here, \texttt{\{objects\}} contains the object labels detected
by YOLO. Given these labels and the previously generated caption,
we prompt the captioning model to assign matching entities
to \texttt{yolo} and the remaining entities to \texttt{caption},
treating synonyms as matches and using YOLO labels for them.
YOLO labels are provided only for this classification step,
not during initial caption generation.


\section{Statistical Comparison}
\label{app:statistical_comparison}
\begin{table}[!t]
\centering
\caption{\textbf{Statistical comparisons on 3B model.}
$\Delta$ is ReMo minus each comparator in percentage points. Ablations report mean $\pm$ standard deviation of paired differences over five seeds. CIs and $p$-values use paired bootstrap over items.}
\label{tab:statistical_comparison}
\label{tab:statistical_ablation}
\vspace{-6pt}
\setlength{\tabcolsep}{5pt}
\scriptsize
\resizebox{\linewidth}{!}{%
\begin{tabular}{lccc}
\toprule
Method & $\Delta$ & 95\% CI & $p$ \\
\midrule
\multicolumn{4}{c}{Overall} \\
\midrule
Full Token & $+0.16$ & $[-0.59,+0.91]$ & $0.69$ \\
DASH & $+1.32$ & $[+0.59,+2.07]$ & $<0.001$ \\
\midrule
\multicolumn{4}{c}{World.} \\
\midrule
w/o local & $+0.88\pm0.85$ & $[+0.20,+1.58]$ & $0.0090$ \\
w/o text & $+0.85\pm0.74$ & $[+0.30,+1.45]$ & $0.0017$ \\
w/o rope & $+0.73\pm0.45$ & $[+0.15,+1.33]$ & $0.013$ \\
w/o proj. & $+0.88\pm0.63$ & $[+0.28,+1.50]$ & $0.0048$ \\
w/o temporal & $+0.05\pm0.47$ & $[-0.33,+0.43]$ & $0.84$ \\
\bottomrule
\end{tabular}%
}
\vspace{-6pt}
\end{table}
We evaluate Qwen2.5-Omni$_{3\mathrm{B}}$, sampling with temperature $0.3$ and top-$p$ $0.9$. For each evaluation item, we average correctness across seeds and compute the paired difference between methods. We obtain 95\% confidence intervals and $p$-values by paired bootstrap resampling over items. The comparisons with Full Token and DASH combine evaluation items from World., Daily., MME, and OmniVid. The component ablations use the same procedure on World.

\subpara{Comparison with Full Token and DASH.}
We select DASH as the strongest compression baseline on 3B under greedy decoding in \cref{tab:main}.
\cref{tab:statistical_comparison} shows comparable performance over five seeds (0--4) to Full Token, with an accuracy difference of $+0.16$ percentage points (95\% CI $[-0.59,+0.91]$, $p=0.69$). ReMo improves accuracy over DASH by $1.32$ points (95\% CI $[+0.59,+2.07]$, $p<0.001$).

\subpara{Component ablation.}
\cref{tab:statistical_ablation} reports the mean and standard deviation of paired accuracy differences across five seeds, together with bootstrap confidence intervals and $p$-values. Removing local detail, text proxies, text position, or projection lowers accuracy on World., with all four comparisons significant at the nominal $0.05$ level. Removing text proxies lowers accuracy by $0.85$ points, while removing text position lowers it by $0.73$ points even when the proxies remain. These results suggest that both object identities and their temporal and spatial grounding contribute to performance. Temporal merging primarily reduces the input token count: removing it lowers compression in \cref{tab:ablation}, while leaving mean accuracy on World. nearly unchanged ($p=0.84$).

\section{Qualitative Analysis}
\label{app:qualitative}
We illustrate how text proxies complement the retained visual evidence, where they can mislead the model, and how the common-space projection changes pruning and retrieval. The examples show both the benefits and the limits of redistributing visual information across modalities.

\subsection{When Text Proxies Help}
\label{app:proxy_success}
\cref{fig:proxy_case} shows three questions for which ReMo and the full-token model answer correctly, while OmniSelect does not. The injected object names provide compact semantic anchors alongside the retained visual tokens: \textit{car} in the vehicle demonstration, \textit{person} and \textit{sports ball} in the exercise video, and \textit{person} and \textit{bird} in the outdoor scene. By providing object identities together with their temporal and spatial positions, the proxies can reduce the need to identify and localize the relevant objects from sparse visual tokens. This allows the model to focus on the relations required by the question: associating the rear lettering with the propeller demonstration, comparing the ball's size with the person's height, and connecting the person's reaction to the bird-hunting event. These examples suggest that spatially grounded text proxies help preserve the object context needed for temporal, spatial, and causal reasoning after pruning.

\subsection{When Text Proxies Mislead}
\label{app:proxy_failure}
\cref{fig:failure_case} shows three failure cases in which the full-token model is correct but ReMo is not. In the news example, the \textit{tie} proxy highlights an attribute appearing in an incorrect option, while the correct answer depends on the backgrounds. In the seed-shop example, labels such as \textit{book}, \textit{tv}, and \textit{suitcase} introduce misleading object semantics, and ReMo selects the shop-interior option instead of the event immediately following the tray of plants. In the snowmobile example, YOLO detects only one of the three vehicles, so the injected \textit{car} proxies convey an incomplete set of detections. ReMo consequently answers one instead of three. These cases illustrate how text proxies can emphasize distracting attributes or propagate detector errors, including incorrect labels and missed objects, into downstream reasoning.

\subsection{Common Space Alignment}
\label{app:alignment_cases}
\cref{fig:ccaw_case} illustrates the common-space projection in \cref{sec:proj}. In the upper two examples, projection removes additional visual tokens around the handled object producing mechanical clicks and the speaking news anchor. These regions correspond to the sound sources, consistent with the intended role of audio saliency: identifying visual content already represented by audio. Projection changes which visual tokens are pruned; the audio stream is retained.

The lower two examples show audio-to-video retrieval. For the snare-drum and table-tennis queries, the paired video's rank improves from 46 and 89 in the raw space to 1 in the common space. The other top-ranked videos also become semantically related to the query, showing percussion performances and racket sports, respectively. These examples complement the retrieval results by showing how alignment makes cross-modal similarity more informative for both matching videos and identifying audio-explainable visual regions.

\subpara{Clip-level calibration and patch-level saliency.}
Although the CCA projections are estimated from clip-level audio--visual pairs, the visual projection is applied separately to each patch at inference. Each projected patch is compared with the chunk's mean audio representation using the audio saliency in \cref{eq:saliency_terms}. The shared audio anchor therefore does not assign a uniform score to the frame: patches that agree with the audio receive lower audio saliency, while those carrying complementary visual information receive higher scores. The upper examples in \cref{fig:ccaw_case} provide qualitative evidence that this calibration can produce spatially selective pruning around sound sources without object-level correspondence supervision.

A related observation appears in LiteLVLM~\cite{lee2026cliptricks}, which scores visual patches against a global CLIP text embedding and retains low-similarity patches for pixel grounding. This provides a precedent for retaining informative local tokens according to their dissimilarity to a global cross-modal reference, supporting the scoring choice in \cref{eq:saliency_terms}. Together with the spatially selective pruning in \cref{fig:ccaw_case}, it motivates applying the clip-calibrated projection to individual visual patches for saliency estimation.

\begin{figure*}[p]
    \centering
    \includegraphics[width=\textwidth,height=0.84\textheight,keepaspectratio]{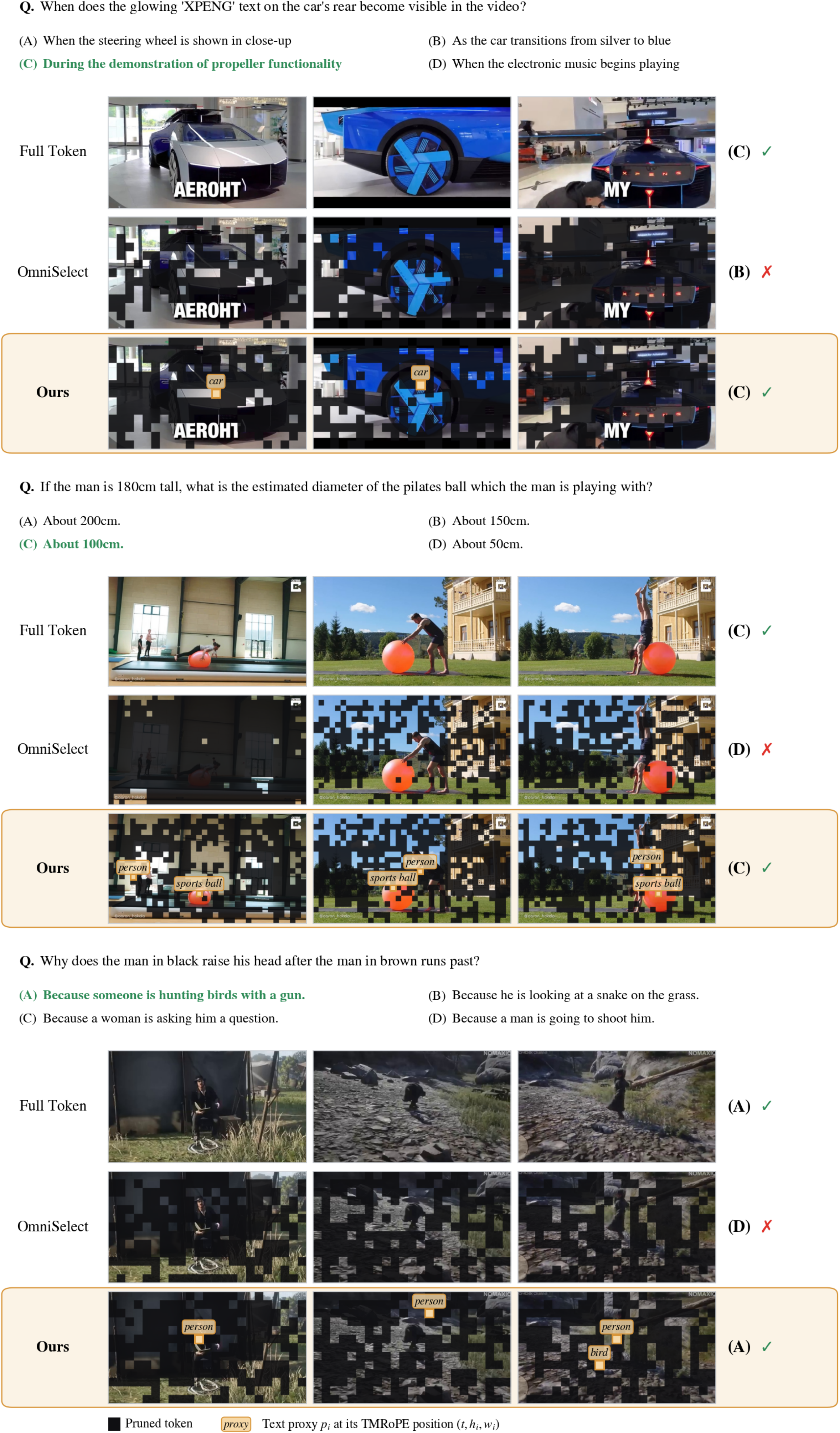}
    \caption{\textbf{Text proxies supporting question answering after pruning.} Three examples compare the full-token model, OmniSelect, and ReMo. ReMo retains the correct answers while exposing object identities through spatially grounded proxies. Black patches indicate pruned visual tokens, and orange labels show text proxies at their TMRoPE positions. By providing object identities and locations, the proxies help the model focus on temporal, spatial, and causal relations in the retained evidence.}
    \label{fig:proxy_case}
\end{figure*}

\begin{figure*}[p]
    \centering
    \includegraphics[width=\textwidth,height=0.84\textheight,keepaspectratio]{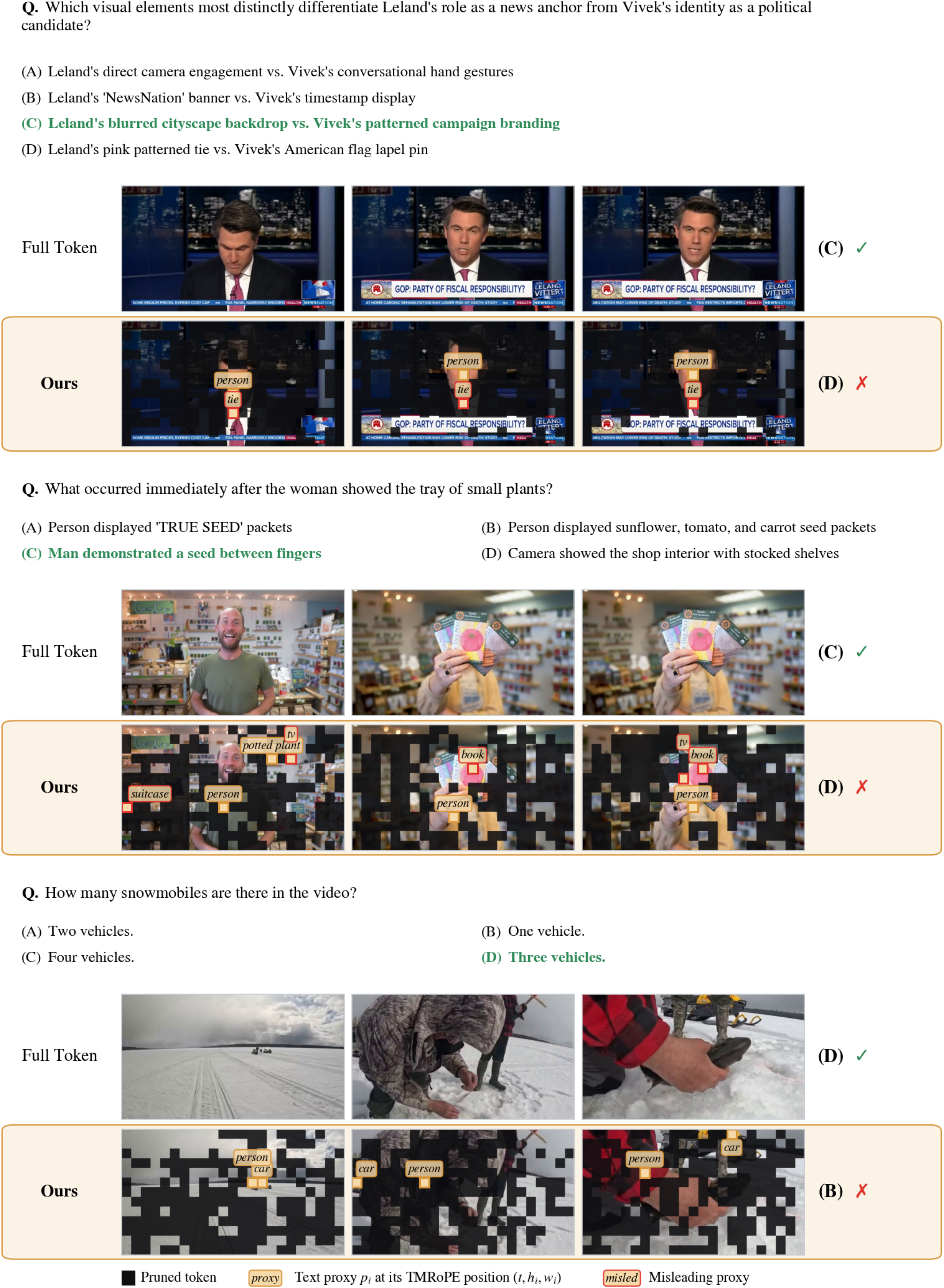}
    \caption{\textbf{Failure cases of semantic offloading.} ReMo answers incorrectly where the full-token model succeeds. From top to bottom, the examples illustrate emphasis on a distracting attribute, incorrect detector labels, and a missed-detection error in which YOLO detects only one of three vehicles. Black patches indicate pruned visual tokens, orange labels denote text proxies, and red outlines highlight misleading proxies. The examples show how distracting labels and detector errors can mislead downstream reasoning.}
    \label{fig:failure_case}
\end{figure*}

\begin{figure*}[p]
    \centering
    \includegraphics[width=\textwidth,height=0.84\textheight,keepaspectratio]{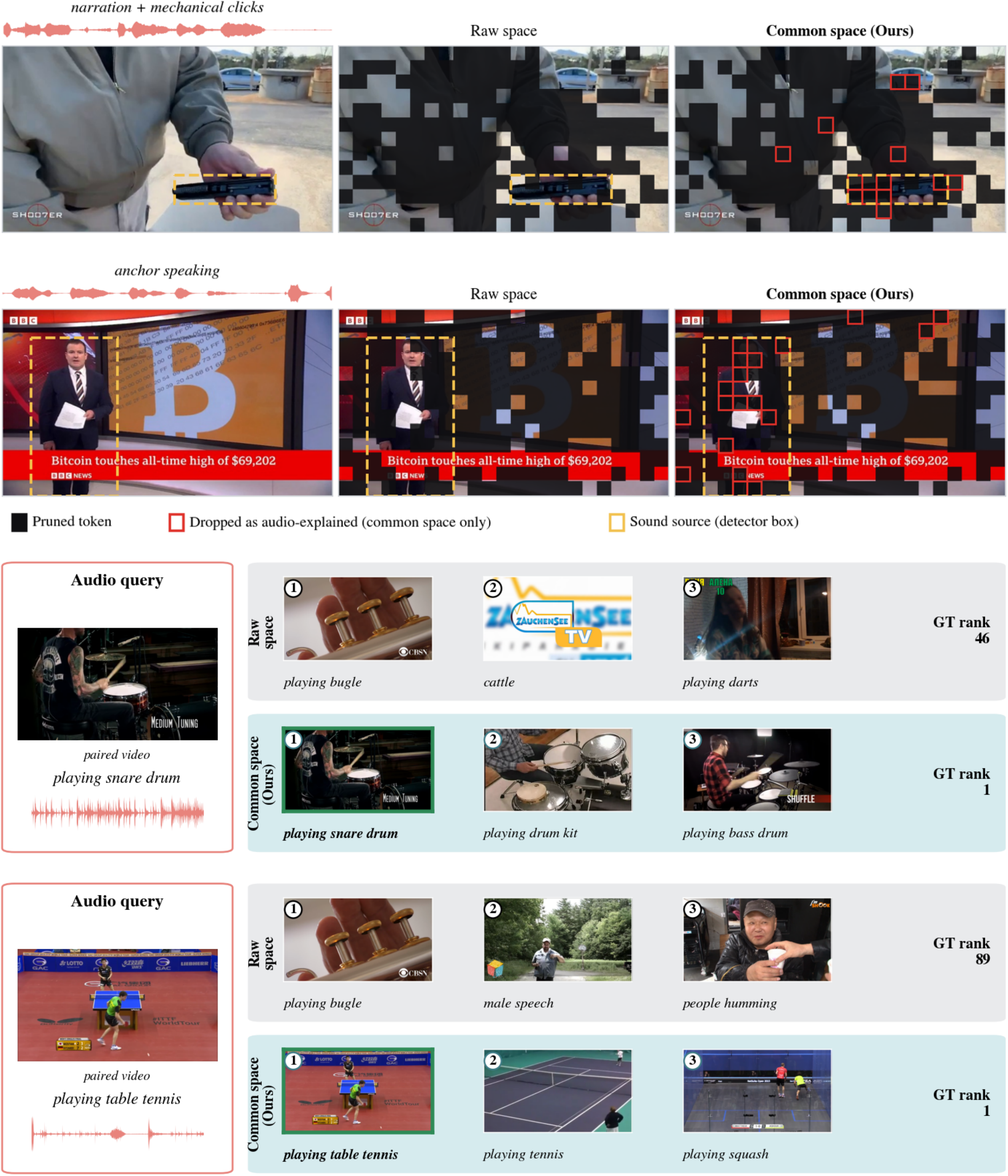}
    \caption{\textbf{Common space alignment for visual pruning and audio-to-video retrieval.} Top: compared with the raw space, projection prunes additional tokens around sound sources. Yellow dashed boxes mark sound sources, red outlines mark tokens dropped only in the common space, and black patches indicate pruned tokens. Bottom: the paired videos for snare-drum and table-tennis audio move from ranks 46 and 89 to rank 1; green borders mark the retrieved paired videos.}
    \label{fig:ccaw_case}
\end{figure*}

\clearpage

{
    \small
    \bibliographystyle{ieeenat_fullname}
    \bibliography{shortstrings,main}
}

\else

\fi
\end{document}